\documentclass{article}
\usepackage{spconf,amsmath,graphicx,subcaption,cleveref,booktabs, multirow}

\title{Eyes on Target: Gaze-Aware Object Detection in Egocentric Video}
\name{Vishakha Lall, Yisi Liu \sthanks{Thanks to Singapore Maritime Institute (SMI)}}
\address{Singapore Polytechnic\\
	Centre of Excellence in Maritime Safety\\
	Singapore}
\begin{document}
\ninept
\maketitle
\begin{abstract}
Human gaze offers rich supervisory signals for understanding visual attention in complex visual environments. In this paper, we propose Eyes on Target, a novel depth-aware and gaze-guided object detection framework designed for egocentric videos. Our approach injects gaze-derived features into the attention mechanism of a Vision Transformer (ViT), effectively biasing spatial feature selection toward human-attended regions. Unlike traditional object detectors that treat all regions equally, our method emphasises viewer-prioritised areas to enhance object detection. We validate our method on an egocentric simulator dataset where human visual attention is critical for task assessment, illustrating its potential in evaluating human performance in simulation scenarios. We evaluate the effectiveness of our gaze-integrated model through extensive experiments and ablation studies, demonstrating consistent gains in detection accuracy over gaze-agnostic baselines on both the custom simulator dataset and public benchmarks, including Ego4D Ego-Motion and Ego-CH-Gaze datasets. To interpret model behaviour, we also introduce a gaze-aware attention head importance metric, revealing how gaze cues modulate transformer attention dynamics.
\end{abstract}
\section{Introduction}
\label{sec:intro}

Wearable eye trackers capture egocentric videos enriched with gaze data, offering unique insights into user attention and intention. As immersive, first-person experiences grow central to domains such as virtual reality (VR), augmented reality (AR), robotics, and simulation-based training, understanding where users look and what they attend to has become increasingly critical. In particular, simulation environments for skill assessment, such as maritime or aviation training, rely on tracking a subject’s visual attention to determine whether they focus on task-relevant objects at the right time. However, traditional object detection models are not designed to leverage gaze signals, limiting their utility in attention-sensitive applications. Incorporating gaze information into object detectors can enable more accurate and human-aligned identification of regions of interest, improving both detection performance and interpretability in egocentric scenarios.

Gaze serves as a critical behavioural signal, reflecting both the spatial distribution of human attention and the underlying cognitive processing of visual scenes \cite{gazestrategy}. Recent research has explored the integration of human gaze into computer vision tasks. The authors in \cite{gazeclassification2, gazeclassification} explore gaze embeddings to use human gaze data for image classification. The authors in \cite{cropping} present an interactive cropping technique that uses eye-tracking data to identify important image content. The work in \cite{gazemedical, gazemedical2} demonstrates the effectiveness of integrating expert human gaze to bias a Vision Graph Neural Network (GNN) in regions of high relevance. In egocentric contexts, gaze patterns offer rich supervisory cues, as described in \cite{videoattention}, which leverages human attention bias to extract semantically significant objects from videos using eye-tracking data. While prior work has leveraged gaze features or used computer vision to estimate gaze, to the best of our knowledge, no existing models incorporate gaze as a contextual input within an object detection framework. In a similar direction, \cite{weaklysupervisedgaze} addresses weakly supervised attended object detection in cultural sites, showing that gaze can substitute for costly manual annotations by leveraging frame-level labels. Complementing this, \cite{asnet} proposes the Attentive Saliency Network (ASNet), which uses fixation maps to guide salient object detection, thereby bridging fixation prediction and object-level segmentation.

Building on this foundation, we explore the alignment between human visual attention and the attention mechanisms used in modern deep learning models, particularly transformers, which are intuitively suited for this task due to their architecture that explicitly models relationships within the input. As originally demonstrated in natural language processing, attention mechanisms were designed to highlight semantically important words within a sequence \cite{vaswani}. We attempt to extend this concept to object detection, where attention layers allow models to prioritise spatial regions that are most relevant for tasks using the gaze data. Recent studies \cite{humanattention} have shown that aligning artificial attention with human attention not only improves task performance but also enhances model interpretability in neural network design. Motivated by this, we investigate how integrating human gaze, an explicit signal of visual attention, can guide a Vision Transformer (ViT) detector to focus on task-relevant objects in an egocentric video. 

In related work, \cite{gazeandattentionclassification} explores the integration of human gaze into spatio-temporal attention mechanisms for the recognition of egocentric activity. While their aim appears similar, their approach fundamentally differs by focusing on scenarios without explicit gaze data. They model gaze fixation locations as discrete latent variables, using a variational method to learn gaze distributions during training and predict gaze during testing. These predicted gaze locations serve as attentional cues, contrasting with our methodology, which directly leverages ground-truth gaze data for alignment and interpretability.
We make two primary contributions:
\begin{enumerate}
    \item We introduce `Eyes on Target', a depth-aware, gaze-guided ViT model for object detection in egocentric videos, where we use the gaze-derived features as a guiding signal to bias attention heads to align with human visual attention patterns and show that it surpasses the performance of contemporary object detection models.   
    \item We build upon the existing `head importance' metric and propose a new metric, `gaze-aware head importance', to quantify the impact of gaze data on attention heads, offering insights into how different gaze integration methods influence attention strength.
\end{enumerate}

\section{Background}
\label{sec:background}

\subsection{Eye Tracking}
Eye tracking technology measures eye positions and movements to identify where a subject is looking. Wearable eye trackers use scene cameras, infrared illuminators, and pupil/cornea reflections to estimate gaze movement in 3D space. Key features relevant to our study include: gaze position (the horizontal and vertical coordinates of gaze on the scene frame), gaze depth ( distance to the object in focus), pupil diameter (an indicator of cognitive attention changes \cite{pupildilation}), and gaze direction (the orientation of the eye movement relative to the position of the head).

\subsection{Attention in ViT}
Self-attention in ViT enables each patch in an image to attend to all others, capturing both local details and global context. ViTs process images by dividing them into fixed-size, non-overlapping patches, which are embedded into high-dimensional vectors. At each self-attention layer, the model generates attention maps that highlight salient patches, emphasising regions critical for object recognition. The attention between image patches $i$ and $j$ in a ViT is computed using the dot product of their query $q_i$ and key $k_j$ vectors, followed by a softmax operation to normalise the attention weights. This is mathematically expressed as:
\begin{equation}
    A_{i,j} = \frac{q_i \cdot k_j}{\sqrt{d_k}} 
\label{eq:attention}
\end{equation} where $d_k$ is the dimensionality of the key vectors, used for scaling to prevent large values. Images are presented to the model as sequences of fixed-size patches (16x16 resolution), which are linearly embedded. A [CLS] token is prepended to the sequence for the label, and absolute position embeddings are added for bounding boxes.

\subsection{Attention Head Importance Metric}
Recent research has focused on interpreting the internal attention mechanisms of Vision Transformers (ViTs). The work in \cite{doesattentionworkvision} introduces methods for visual analytics that provide insights into the importance of individual attention heads. In ViT models, multiple attention heads in each layer capture different aspects of the input image, generating diverse attention patterns. Head importance measures each head's contribution to local attention distributions and the model's overall performance. Typically, the importance of a head $h$ can be expressed as:
\begin{equation}
I^{prob}_{h} = \frac{1}{N}\sum_{x\in\mathcal{D}}^{}\frac{1}{L}\sum_{j=1}^{L}\text{AttnScore}^{h}_j(x)    
\label{eq:head importance metric}
\end{equation} where $\mathcal{D}$ denotes the dataset of input images, $N$ is the size of the dataset, $L$ is the number of patches in the input image, $\text{AttnScore}^{h}_j(x)$ is the attention score generated by head $h$ for image $x$ around patch $j$. It quantifies how much attention the head allocates to different patches of the input image and contributes to the output for that input and is derived from self-attention weights. Mathematically:
\begin{equation}
\text{AttnScore}^{h}_j(x) = \frac{1}{L}\sum_{i=1}^{L}A_{i,j}^{h}(x)    
\label{eq:attention score}
\end{equation} where , $A_{i,j}^{h}(x)$ represents the attention weight assigned by head $h$ to patch $i$ when attending to the patch $j$ in input $x$. A higher $\text{AttnScore}^{h}_j(x)$ indicates that head $h$ allocates more attention to informative regions $j$.

\section{Datasets}
\label{sec:datasets}

\subsection{Egocentric Maritime Simulator Dataset}
This custom dataset was collected at the Advanced Navigation Research Simulator at Centre of Excellence in Maritime Safety, during maritime training simulations using Tobii Pro Glasses 3. Participants navigated a vessel in the immersive vessel bridge simulator. Participants also interacted with the operational equipment and control panels displaying critical information such as speed, engine status, radar charts, alarms etc. The setup allowed participants to navigate the space freely, enabling them to approach or move away from panels throughout the room. Each participant underwent calibration of the eye tracker, with adjustments made for those requiring prescription lenses by fitting corrective lenses in the eye tracker frame. All data collection involving human participants was conducted with appropriate informed consent and approved by the relevant institutional review board (IRB). Participant data was fully anonymised to ensure privacy and confidentiality. The dataset comprises $\sim$10 hours of egocentric video from 10 different participants, recorded at 30 fps with gaze data sampled at 100 Hz. Around 1 million extracted frames were manually annotated to identify the gaze-focused object (panels and equipment) labels across 36 classes, with bounding boxes marking these target objects. To ensure balanced representation, data validation was performed to verify adequate sample counts within each class distribution. The dataset was then partitioned into training, validation, and test sets following a 70:15:15 split.

To demonstrate the practical utility of the proposed methodology, each video was temporally annotated to indicate the onset and duration of specific high-demand events injected during the simulation exercises, namely, poor visibility, high traffic congestion, and engine failure. These annotated segments correspond to periods of heightened attentional demand and are later used to assess participant performance under stress-critical conditions.

\subsection{Ego Motion Dataset}
To demonstrate the generalisability of our approach, we utilised a publicly available eye-motion dataset captured during desk work activities \cite{egodataset}. The dataset includes over two hours of recordings combining eye-tracking data from an EMR-9 device and high-resolution video from a GoPro HERO 2 HD camera. Synchronisation between devices was achieved using global optical flow alignment, following the method in \cite{egodataset}. The dataset covers five tasks, reading, video watching, text copying, writing, and browsing, performed by five participants for $\sim$2 minutes each, repeated in sequence. Transitions included 30-second void activities to simulate real-world conditions, with variations like different reading materials, video content, and workspace layouts. Each session produced 25–30 minutes of continuous video, contributing to the dataset's diversity and realism. The frames extracted from the dataset were partitioned into training, validation, and test sets following a 70:15:15 split. Since the dataset contains only classification labels and lacks bounding box annotations, it is used exclusively for comparative analysis of classification performance. 

\subsection{Ego-CH-Gaze}
To evaluate our approach, we employ the weakly supervised attended object detection dataset introduced in \cite{weaklysupervisedgaze}. The dataset consists of egocentric videos and gaze coordinates collected from subjects visiting a museum, along with frame-level annotations indicating the class of the attended object among 14 distinct classes. We benchmark our methodology against this dataset.

\subsection{CUB-GHA Dataset}
The CUB-GHA dataset \cite{cubgha} extends the Caltech-UCSD Birds-200-2011 (CUB) dataset with human gaze annotations to support research on attention-guided learning. It contains images of 200 fine-grained bird species, each paired with gaze data collected from five human participants via eye-tracking. Participants were instructed to identify bird species while their fixations were recorded, resulting in gaze heatmaps that highlight discriminative visual regions used in decision-making. These gaze annotations provide ground-truth signals of human visual attention. As the dataset includes only aggregated gaze heatmaps rather than raw gaze feature vectors, it was utilised specifically for parameter tuning of the gaze-aware head importance metric rather than for model evaluation.

\section{Proposed Methodology}
\subsection{Data Preprocessing}
\begin{itemize}
    \item Gaze alignment: The frames and corresponding gaze data are aligned to extract accurate gaze features for each frame.
    \item Gaze Normalisation $g_{x,y}$: The gaze position is normalised to the frame dimensions, converting 2D pixel coordinates into a normalised coordinate system where $(x,y) \in [0,1]$.
    \item Depth Normalisation $d$: Depth values are normalised between $[0,1]$ to the video’s depth scale, assigning $0$ to the nearest objects and $1$ to the farthest, ensuring consistent depth representation across frames of the same video. During inference, the depth scale remains constant for the entire video, ensuring each frame's depth is contextualised within the overall scene for consistent depth interpretation.
    \item Pupil Diameter Normalisation $p$: The pupil diameters from both eyes are normalised to $[0,1]$,  with $0$ representing the smallest observed diameter (minimal attentional focus) and $1$ the largest (maximum attentional focus) as per \cite{pupildilation}. During inference, the normalisation range is established during calibration by having participants perform activities requiring high attention (e.g., reading) and low attention (e.g., gazing) to account for variations across subjects.
    \item Gaze Direction Encoding $\hat{g}$: The gaze direction vectors from both eyes are combined into a single unit vector, providing a directional reference for attention focus.
    \item Data Augmentation by Gaze Position Shifting: To enhance robustness and variability, we augment the dataset by slightly shifting gaze positions along the gaze direction, scaled by depth. This maintains proximity to the original gaze point, preserving class labels and preventing misclassification.
    \item Data Augmentation by Frame Modifications: To enhance robustness and variability, we apply augmentations like cropping, rotation, and colour jittering, adjusting gaze information consistently. These techniques simulate real-life first-person perspectives, such as changes in proximity or head direction.
    \item Frame Resizing: The frame is resized to the model's input size, and the normalised gaze position is adjusted accordingly, followed by re-normalisation.
\end{itemize}

\subsection{Eyes on Target}

The proposed detection framework builds upon the DETR (DEtection TRansformer) architecture \cite{model}, which reformulates object detection as a direct set prediction problem using transformer-based encoder-decoder layers. DETR employs a convolutional backbone—commonly ResNet-50 or ResNet-101—pretrained on ImageNet-1k \cite{imagenet} to extract image features, which are then flattened and passed into a transformer encoder. A fixed set of learnable object queries is used in the decoder to predict object classes and corresponding bounding boxes. The standard DETR architecture is adapted to integrate gaze-based inputs, which influence the attention mechanism.

$A_{i,j}$ as defined in \cref{eq:attention} denotes the original attention score between patch $i$ and patch $j$. The modified attention score, $A_{i,j}^\text{gaze}$, is defined as:

\begin{equation}
A_{i,j}^\text{gaze} = A_{i,j} + \alpha \cdot \text{GazeBias}(i,j)
\label{eq modified attention}
\end{equation} where $\alpha$ is a hyperparameter that controls the impact of gaze information, and $\text{GazeBias}(i,j)$ calculates a bias based on the proximity of patch $j$ to the gaze point. This function incorporates the 2D gaze position and direction and is defined as:

\begin{equation}
\text{GazeBias}(i,j) = \frac{1}{1 + ||g_{x,y} - \text{Patch}_j||^2} \cdot (1 + ||\hat{g} \cdot \text{Patch}_j||)
\label{eq bias}
\end{equation}

\begin{figure}
  \centering
  \begin{subfigure}{0.48\linewidth}
    \includegraphics[width= \linewidth]{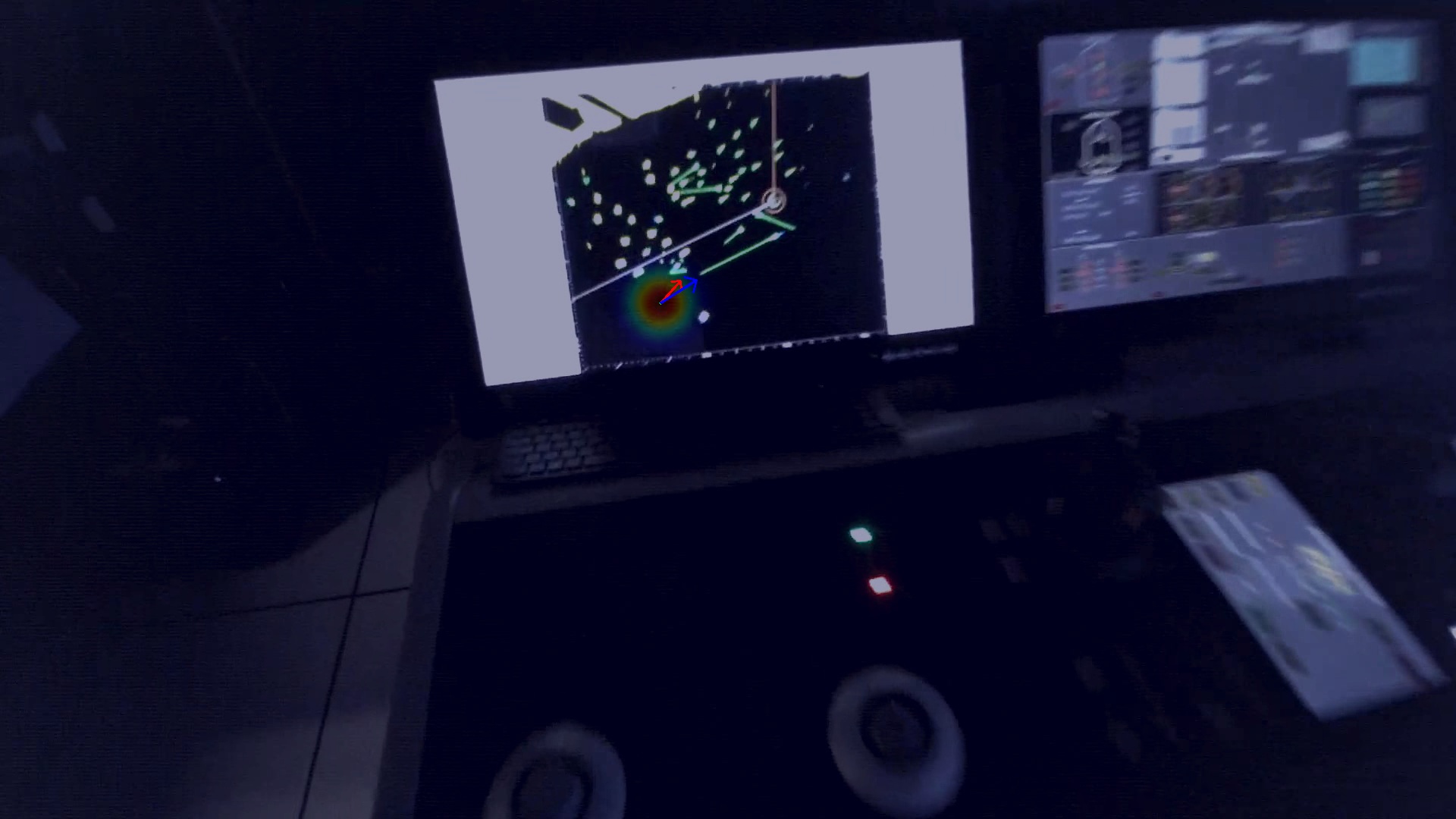}\hfill
    \caption{}
    \label{fig:gazepoint}
  \end{subfigure}
  \hfill
  \begin{subfigure}{0.48\linewidth}
    \includegraphics[width= \linewidth]{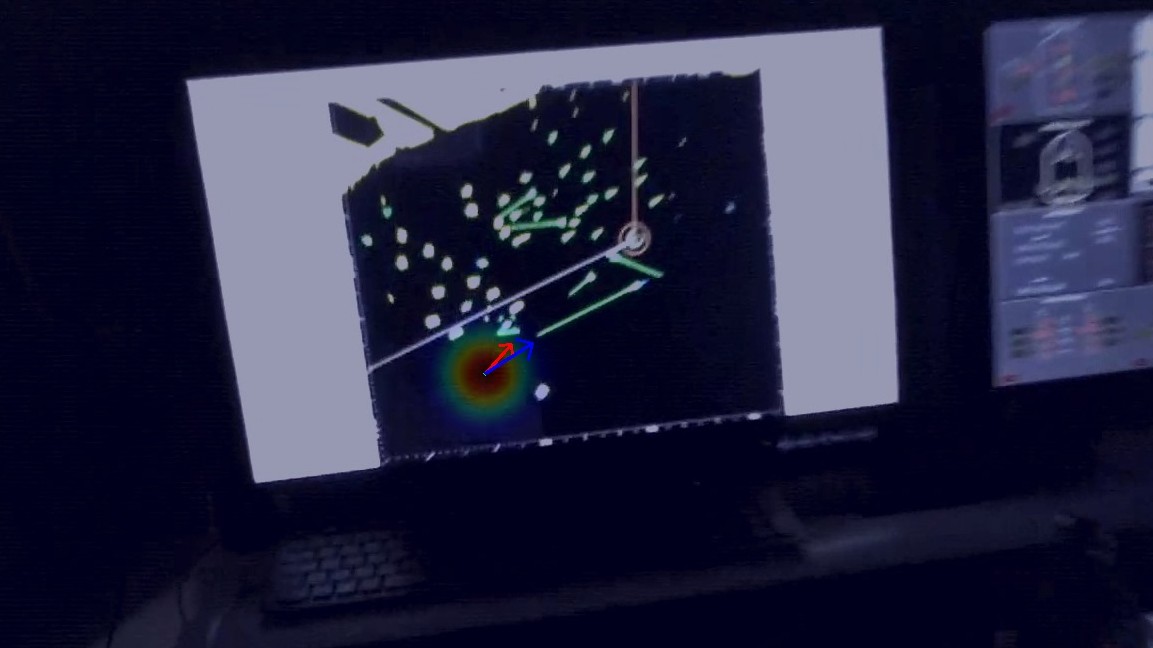}\hfill
    \caption{}
    \label{fig:zoom gazepoint}
  \end{subfigure}
  \begin{subfigure}{0.48\linewidth}
    \includegraphics[width= \linewidth]{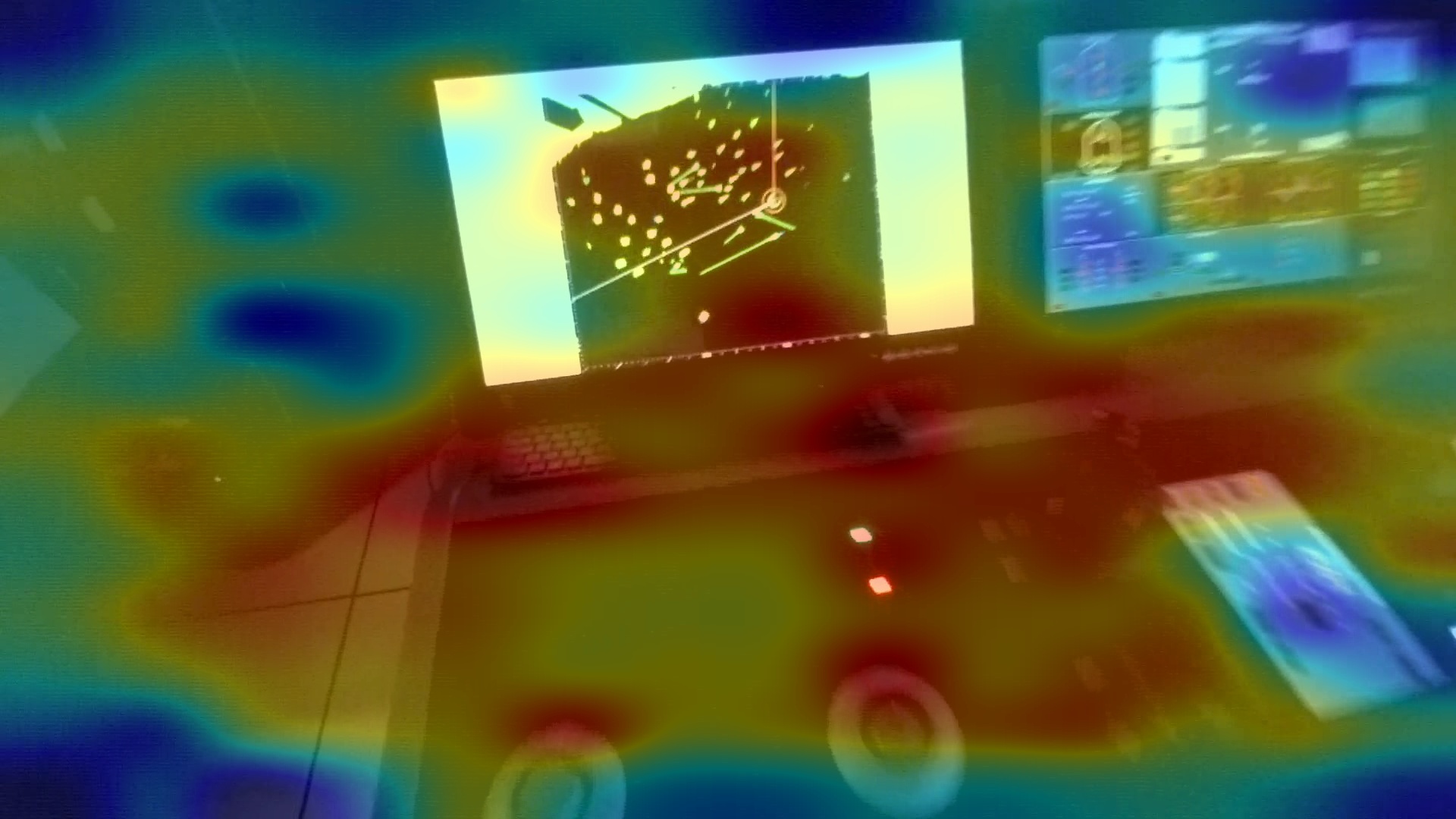}\hfill
    \caption{}
    \label{fig:before attention}
  \end{subfigure}
  \hfill
  \begin{subfigure}{0.48\linewidth}
    \includegraphics[width= \linewidth]{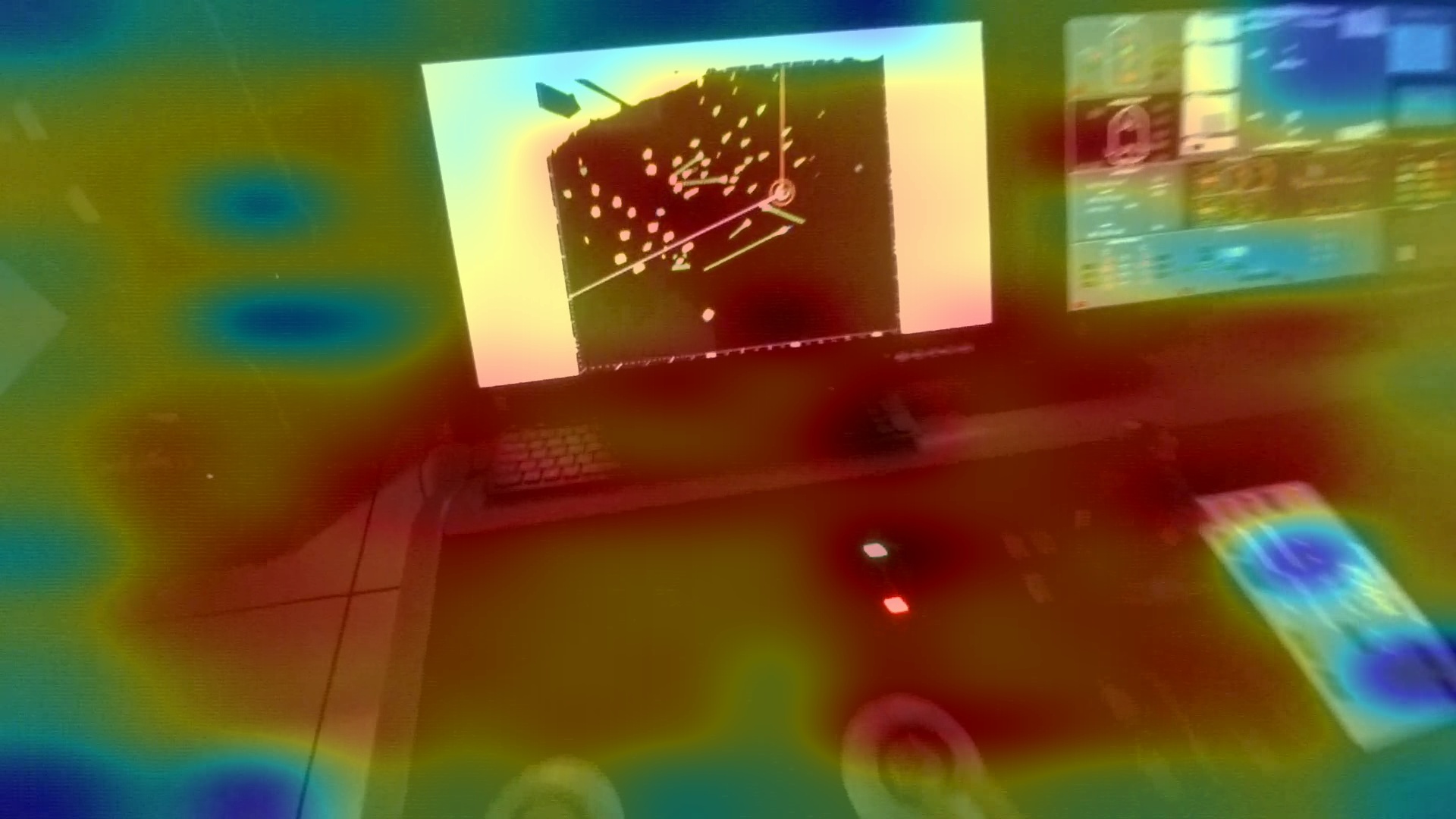}\hfill
    \caption{}
    \label{fig:after attention}
  \end{subfigure}
  \hfill
  \caption{Visualisation of attention modification: (a) Gaze position (circle) and direction (arrows from left and right eye) (b) Zoomed in view of the gaze point (c) Attention map from the original model (d) Attention map after gaze modifications to the original model, showing increased attention in regions around the gaze point.}
  \label{fig:gaze modified attention}
\end{figure}

This formulation ensures that patches near the gaze point, aligned with its direction, receive increased attention weights, enhancing focus on these regions during training. \cref{fig:gaze modified attention} visualises attention modifications using heatmaps rendered in \verb|cv::COLORMAP_JET| style, highlighting enhanced attention around the gaze point and direction, qualitatively confirming the approach's effectiveness.

After computing gaze-modified self-attention maps as described in Eq. \ref{eq modified attention}, the output patch embeddings capture both global scene context and human-relevant focus regions. These enriched representations are passed to a transformer decoder, which predicts object instances by attending to high-attention patches informed by gaze cues. The decoder outputs are fed into two parallel branches: a classification head that assigns object categories to each query, and a bounding box regression head that predicts normalised box coordinates. To further improve localisation, bounding box predictions are refined using dynamically scaled Regions of Interest (RoIs), using a scaling factor $S$ defined as:

\begin{equation}
    S = \lambda_1 \cdot \text{AttnScore} + \lambda_2 \cdot \frac{1}{p} + \lambda_3 \cdot \frac{1}{d}
\label{eq:scaling for bounding box}
\end{equation} where $\lambda_1$, $\lambda_2$ and $\lambda_3$ are hyperparameters that control the influence of each factor, and $\text{AttnScore}$ represents the average attention score within the RoI. The scaling factor $S$ is inversely proportional to pupil diameter and object depth. This enables the model to generate tighter, context-aware boxes around gaze-salient regions.

Both tasks are trained jointly using DETR’s bipartite matching loss: a linear combination of cross-entropy loss for classification and L1 + GIoU loss for bounding box regression, ensuring end-to-end alignment of gaze-informed features with detection outputs. Since the Ego Motion dataset lacks bounding box annotations, we generate pseudo-bounding boxes centred around the gaze point. A fixed-size bounding box is applied uniformly across all frames to approximate regions of interest. These synthetic boxes are then used as inputs to the detector models to infer class labels, enabling a reasonable proxy for classification evaluation in the absence of ground-truth bounding boxes. Training was conducted on an NVIDIA GeForce RTX 4070 GPU with a batch size of 64. For the Egocentric Maritime Simulator Dataset, the model was trained for 5 epochs, and for the Ego Motion dataset, 4 epochs. The number of epochs for each was determined empirically based on the best-performing evaluation metrics. 

During inference and classification evaluation, the detection with the highest confidence score and its corresponding gaze-scaled bounding box is selected, as it is expected to strongly align with the gaze due to our attention modifications. For detection evaluation, however, the full set of predicted bounding boxes is retained to assess overall detection performance.

\subsection{Gaze-Aware Head Importance Metric}
The metric $I^{prob}_{h}$ in \cref{eq:head importance metric} evaluates an attention head's contribution by analysing its attention allocation and consistency across inputs, serving as a baseline to identify influential heads. Gaze data, which indicates where human attention is focused, modifies the head importance metric to emphasise heads aligning with the human gaze. The gaze alignment weight $w^{gaze}_{h}(x)$ measures this alignment for head $h$ on image $x$:

\begin{equation}
    w^{gaze}_{h}(x) = \sum_{i}^{}\text{AttnScore}_{i}^{h}(x)\cdot G(i,x)
\label{eq:gaze alignment weight}
\end{equation}
where $\text{AttnScore}_{i}^{h}(x)$ is the attention score of head $h$ on patch $i$ of image $x$, and $G(i,x)$ is a binary weight indicating the presence of gaze RoI on patch $i$ (where $G(i,x)=1$ if the RoI overlaps with the patch). The final Gaze-Aware Head Importance Metric $I^{gaze}_{h}$ combines the traditional head importance with gaze alignment:

\begin{equation}
    I^{gaze}_{h} = \beta \cdot I^{prob}_{h} + \gamma \cdot \frac{1}{N} \sum_{x\in\mathcal{D}}^{}w^{gaze}_{h}(x)
\label{eq:gaze-aware head importance metric}
\end{equation} where $\beta$ and $\gamma$ are coefficients balancing the baseline importance and gaze alignment. The term $\frac{1}{N} \sum_{x\in\mathcal{D}}^{}w^{gaze}_{h}(x)$ represents the average gaze alignment for head $h$ across the dataset.
In order to quantify changes in head importance as gaze data is incorporated, it is necessary to fine-tune $\beta$ and $\gamma$. The following methodology was used for parameter tuning:

\begin{itemize} 
    \item Dataset: CUB-GHA dataset, which labels human attention on images. 
    \item Baseline: Initially, we set $\beta = 1.0$ and $\gamma = 0.0$. 
    \item Parameter Adjustment: We iteratively adjusted $\beta$ and $\gamma$, tracking the Intersection over Union (IoU) between gaze-aligned attention heatmaps $I^{gaze}_{h}$ and human attention.
    \item Optimal Selection: The optimal parameters ($\beta = 0.7, \gamma=0.3$) from \cref{tab:alpha beta selection} were chosen based on the best IoU.
\end{itemize}

\subsection{Evaluation Metrics}

To assess the performance of our gaze-guided object detection model, `Eyes on Target', we use standard metrics from object detection. We report mAP at IoU thresholds of 0.5 and 0.75 (mAP@0.5 and mAP@0.75), following COCO-style evaluation on the Ego-CH-Gaze and Egocentric Maritime Simulator Dataset. This measures both detection accuracy and localisation precision. Model performance is benchmarked against YOLOv7 and DETR, both trained on the same dataset splits. For the Ego Motion dataset, which lacks bounding box annotations, we generate pseudo bounding boxes centered around gaze points with a fixed size, and only evaluate model performance using classification accuracy. All models were trained under the same conditions and tuned until peak performance was observed on the respective validation sets. 

We assess the contribution of each gaze component (position, depth, pupil diameter, and direction) through ablation studies with the following model variants: no gaze component, the standard DETR attention mechanism; including gaze position $g_{x,y}$, modulates attention using only gaze position, excluding other gaze components; including gaze position $g_{x,y}$ and direction $\hat{g}$, integrates gaze position and direction, excluding pupil diameter and depth. These models are compared against the model that includes all gaze components, allowing us to evaluate the effects of integrating each gaze feature on label classification accuracy for the Egocentric Maritime Simulator Dataset.

\section{Results}
 \subsection{Performance Evaluation}
\begin{table}[]
\centering
\begin{tabular}{cccc}
\textbf{$\beta$} & \textbf{$\gamma$} & \textbf{Mean IoU} \\ \toprule
1.0                            & 0.0                                                        & 0.35               \\
0.9                            & 0.1                                                        & 0.43               \\
0.7                            & 0.3                                                       & \textbf{0.71}               \\
0.5                            & 0.5                                                       & 0.67       
\end{tabular}
\caption{Optimal selection of $\beta$ and $\gamma$ for $I^{gaze}_{h}$}
\label{tab:alpha beta selection}
\end{table}

\cref{tab:classification metrics} presents a comparative analysis of performance metrics across contemporary and state-of-the-art models. As the Ego Motion dataset lacks bounding box annotations, we evaluate our gaze-guided detection model by comparing its classification accuracy against the self-reported performance of the original authors, who used a saccade-based method. This comparison aims to investigate the benefits of integrating gaze information and assess its effectiveness relative to non-gaze-aware baselines. It is intuitive that the MoWord+SaWord method \cite{egodataset}, which relies primarily on gaze features such as saccades rather than visual content or inter-object relationships, underperforms in complex environments. While effective for activity recognition tasks in datasets like Ego Motion, focused on actions such as reading or watching videos, it struggles on the Egocentric Maritime Simulator Dataset, where classification depends heavily on the visual scene context rather than gaze dynamics alone. A similar trend is observed on the Ego-CH-Gaze dataset. The Attended Object of Interest Detection baseline \cite{weaklysupervisedgaze} achieves modest classification performance, reflecting the limitations of relying on weak supervision and frame-level labels without strong visual grounding, especially in the Egocentric Maritime Simulator Dataset, which has 36 classes compared to the 14 in Ego-CH-Gaze. Conversely, our proposed Eyes on Target model exhibits substantially higher classification accuracy on all the datasets. This can be attributed to the deep features extracted from regions centred around gaze points, as defined by our pseudo-bounding box strategy. Notably, our gaze-integrated model outperforms existing state-of-the-art baseline methods on these datasets as well. The results reinforce the utility of incorporating gaze as contextual guidance, even in settings with constrained spatial dynamics.

To compute classification accuracy for YOLOv7 and DETR, we assign the label of the detection whose bounding box contains the gaze point, treating it as the predicted class. As hypothesised, our gaze-aware model significantly outperforms these baselines, demonstrating that directly integrating gaze information into the attention mechanism yields better alignment with human visual focus than post-hoc matching based solely on spatial overlap. 

\begin{figure}[h]
  \centering
  \begin{subfigure}{0.3\linewidth}
    \includegraphics[width= \linewidth]{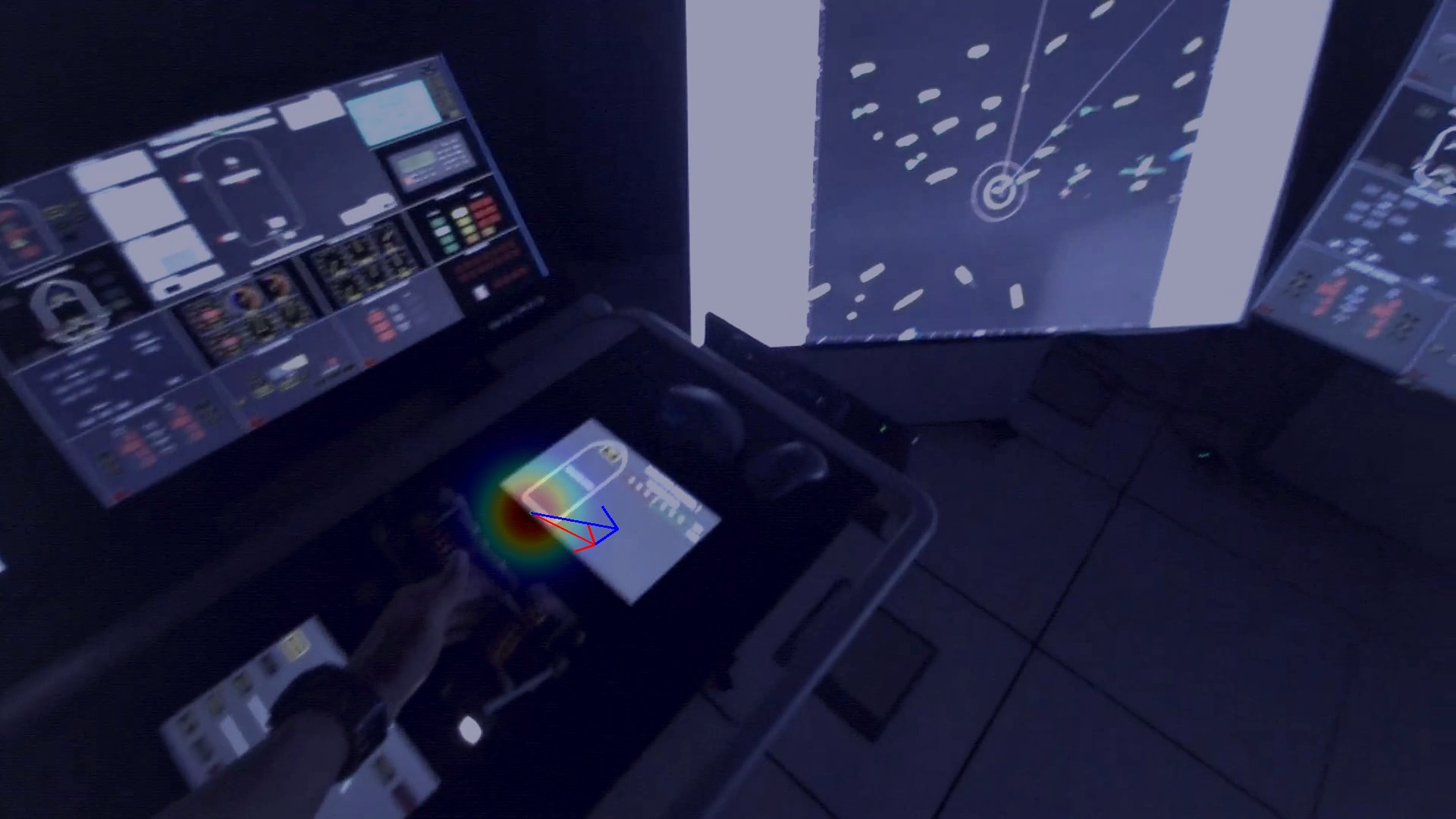}\hfill
    \caption{$d=0.35$}
  \end{subfigure}
  \hfill
  \begin{subfigure}{0.3\linewidth}
    \includegraphics[width= \linewidth]{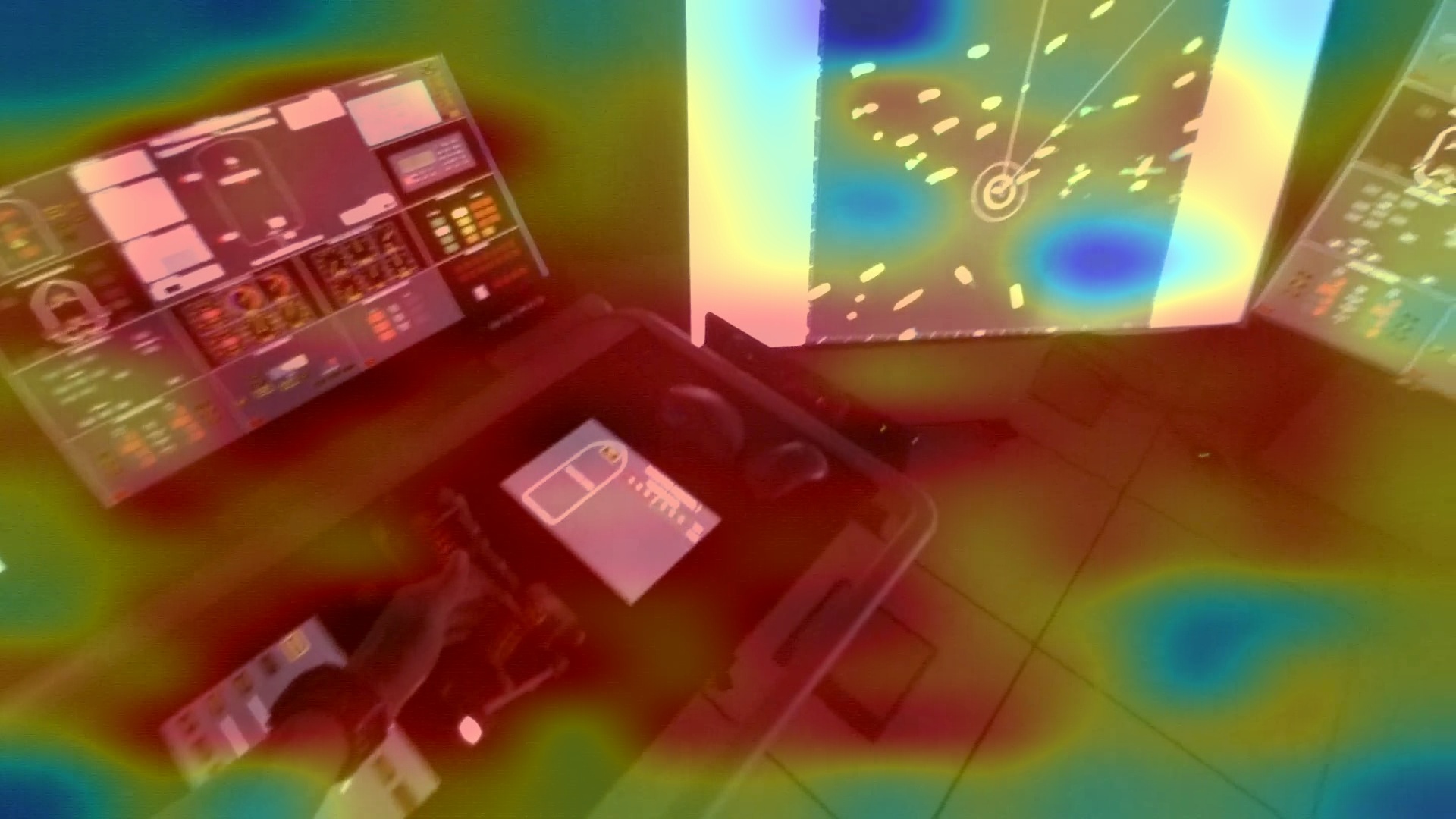}\hfill
    \caption{$p=0.85$}
  \end{subfigure}
  \hfill
  \begin{subfigure}{0.3\linewidth}
    \includegraphics[width= \linewidth]{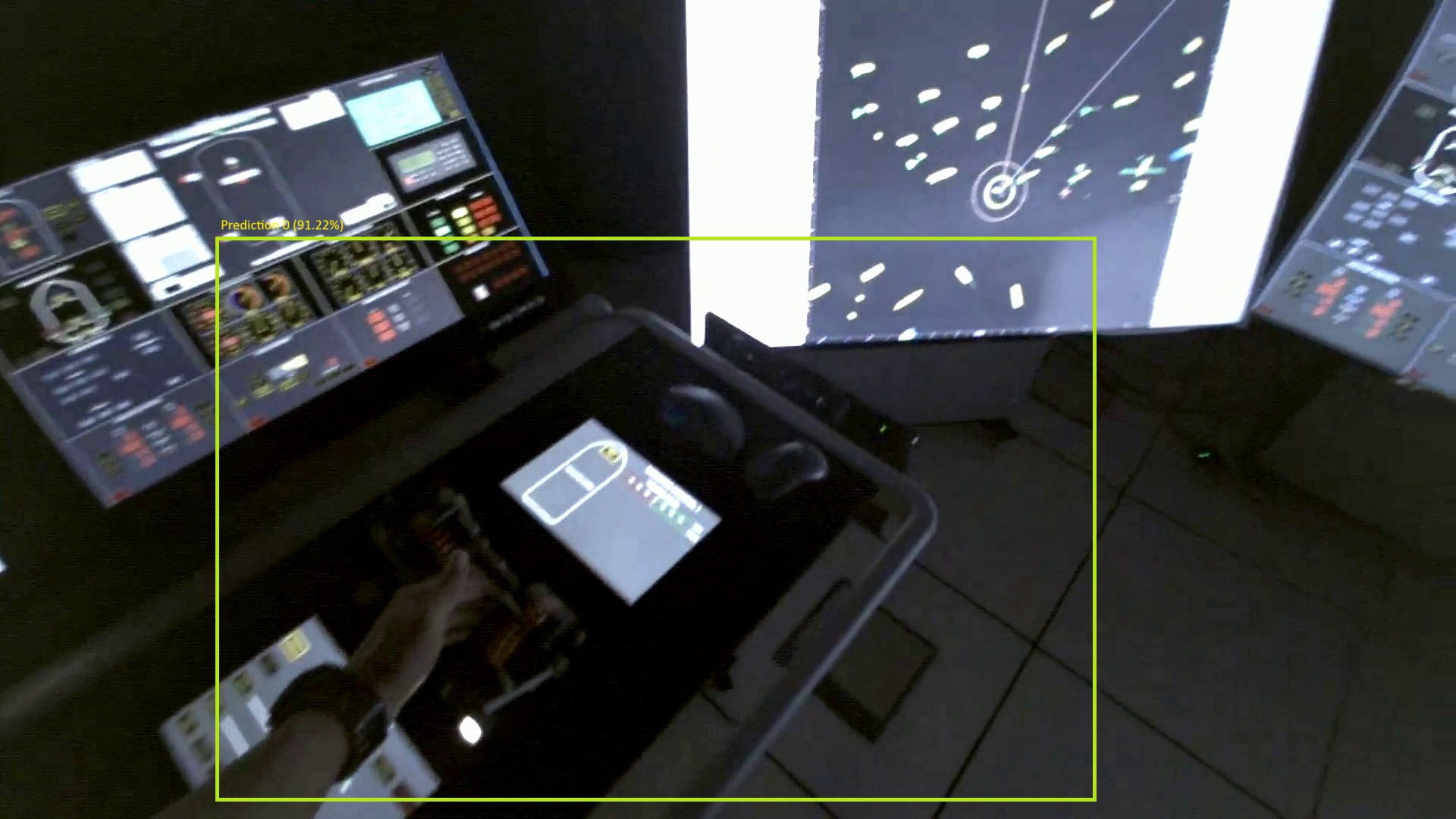}\hfill
    \caption{BowThruster}
  \end{subfigure}
  \hfill
  \begin{subfigure}{0.3\linewidth}
    \includegraphics[width= \linewidth]{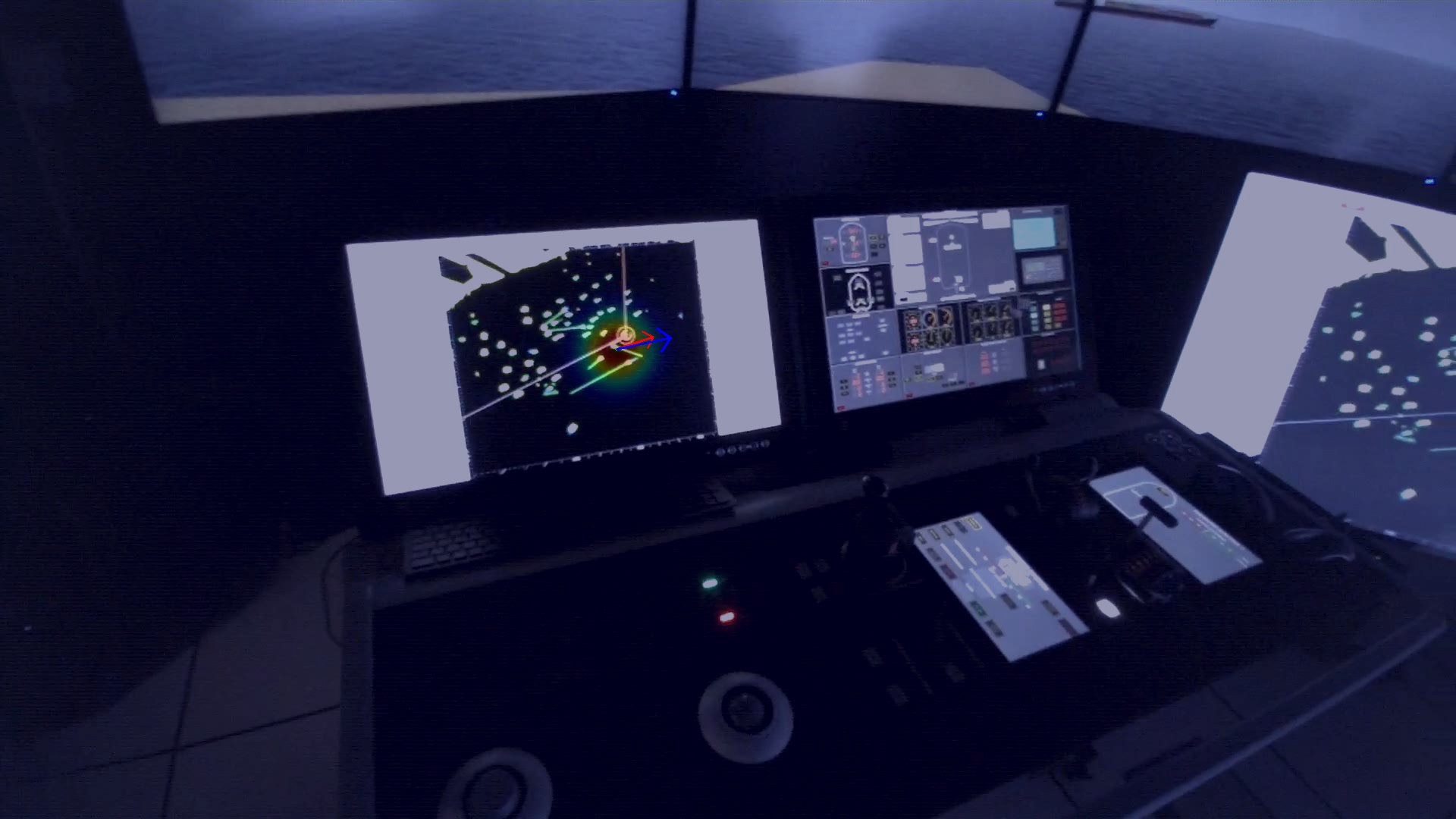}\hfill
    \caption{$d=0.52$}
  \end{subfigure}
  \hfill
  \begin{subfigure}{0.3\linewidth}
    \includegraphics[width= \linewidth]{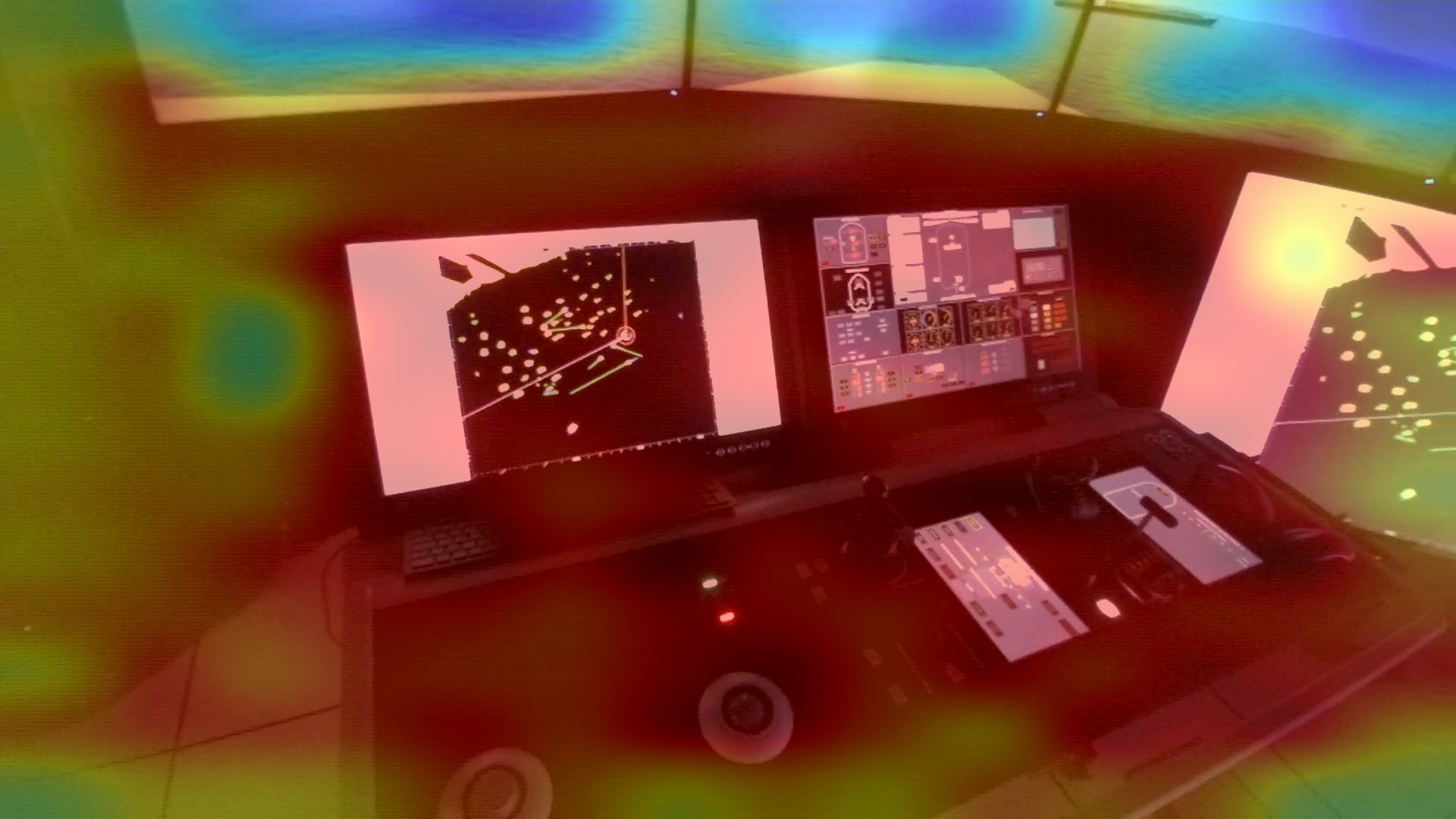}\hfill
    \caption{$p=0.65$}
  \end{subfigure}
  \hfill
  \begin{subfigure}{0.3\linewidth}
    \includegraphics[width= \linewidth]{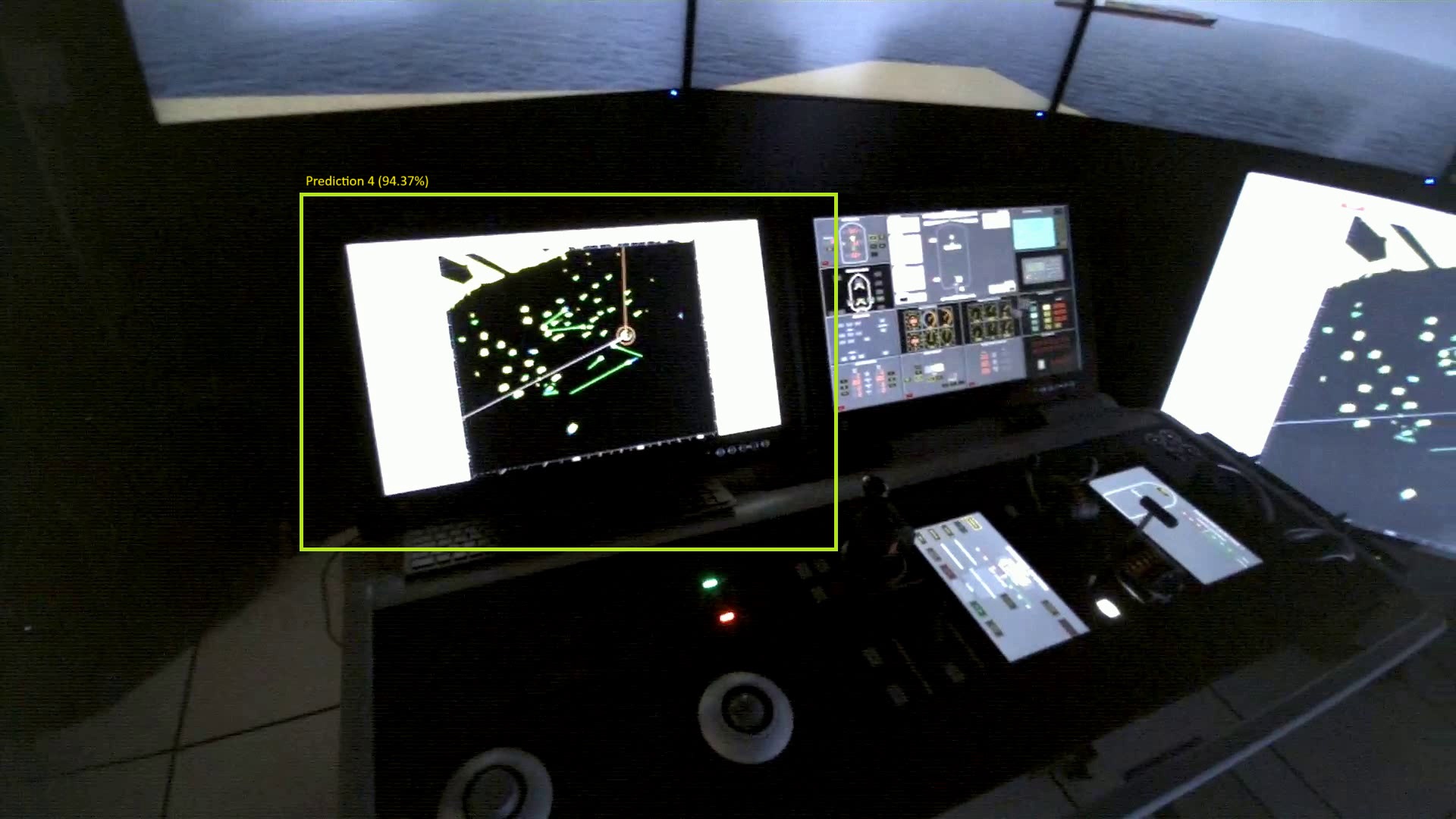}\hfill
    \caption{Radar}
  \end{subfigure}
  \hfill
  \caption{Examples from the Egocentric Maritime Simulator Dataset test set, showing (from left to right) gaze points on original frames with target depth $d$, modified attention heatmaps with pupil dilation $p$, bounding box localisations and classification results, with IoU against ground truth. Frames with closer targets (a:c) display stronger heatmaps and larger boxes due to focus intensity and depth scaling, while having high IoU with ground truth (0.94), while frames distant (d:f) show smaller bounding boxes with strong IoU (0.92)}
  \label{fig:results cems}
\end{figure}

\begin{figure}[h]
  \centering
  \begin{subfigure}{0.3\linewidth}
    \includegraphics[width= \linewidth]{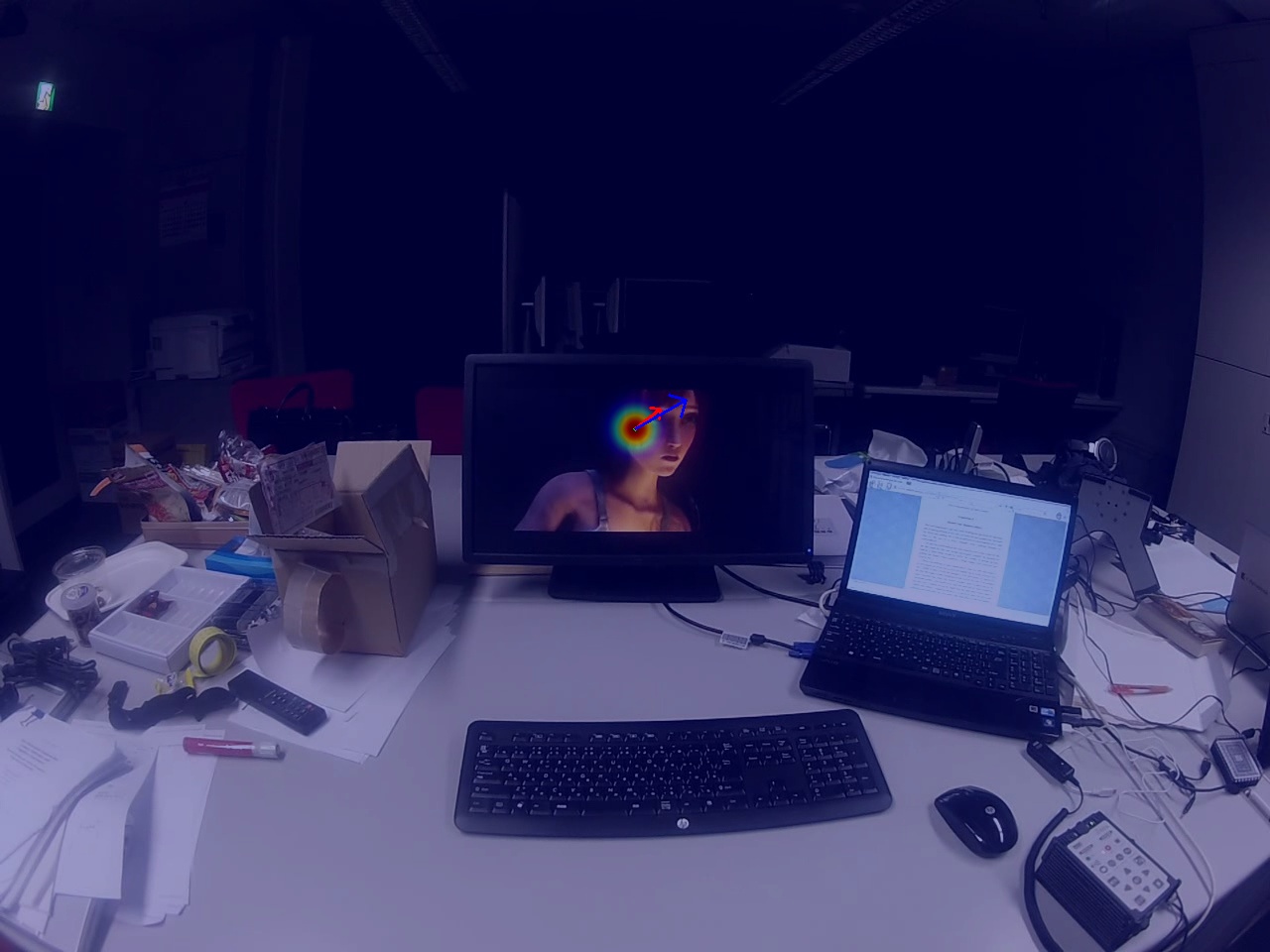}\hfill
    \caption{$d = 0.3$}
  \end{subfigure}
  \hfill
  \begin{subfigure}{0.3\linewidth}
    \includegraphics[width= \linewidth]{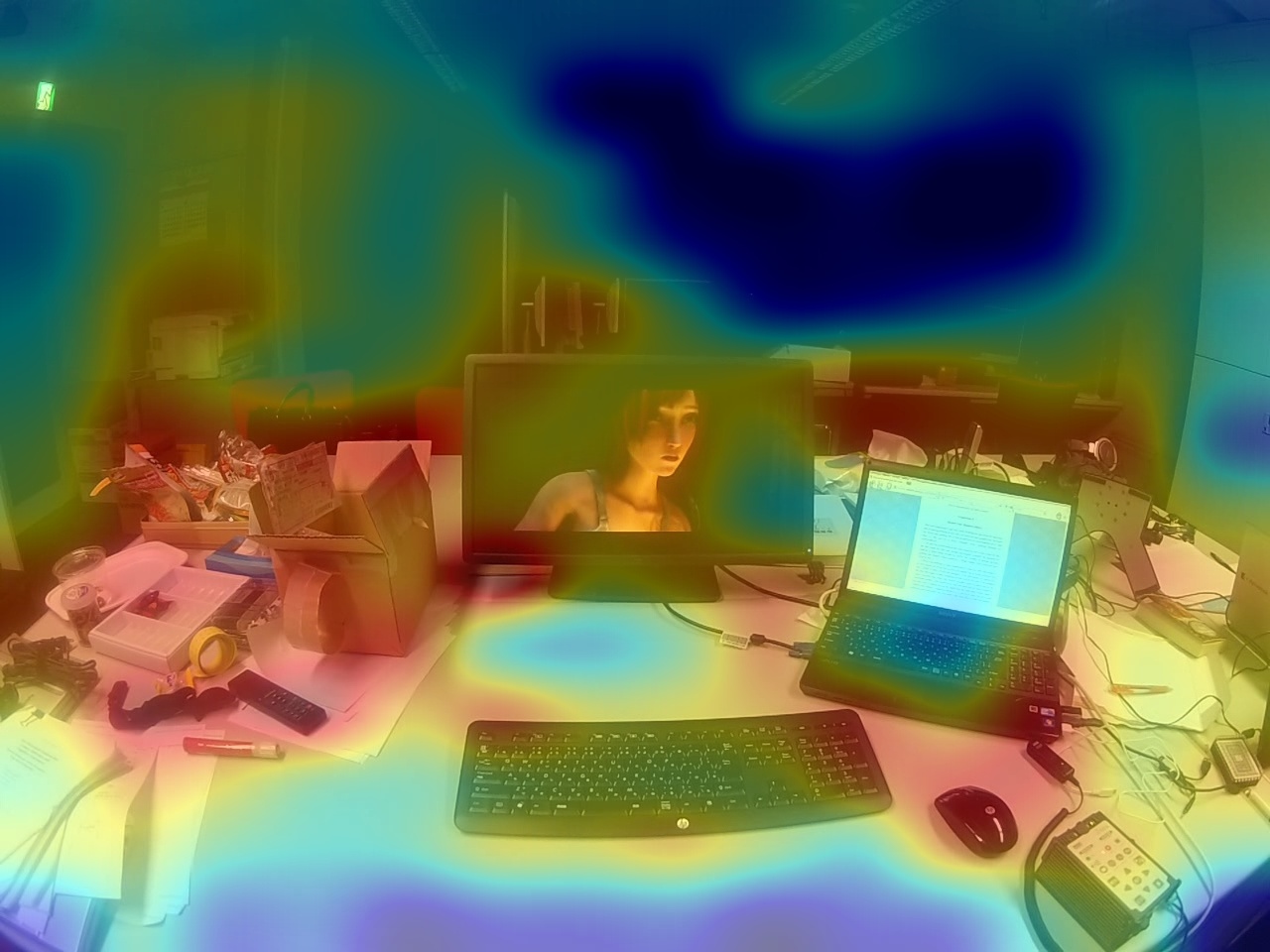}\hfill
    \caption{$p = 0.88$}
  \end{subfigure}
  \hfill
  \begin{subfigure}{0.3\linewidth}
    \includegraphics[width= \linewidth]{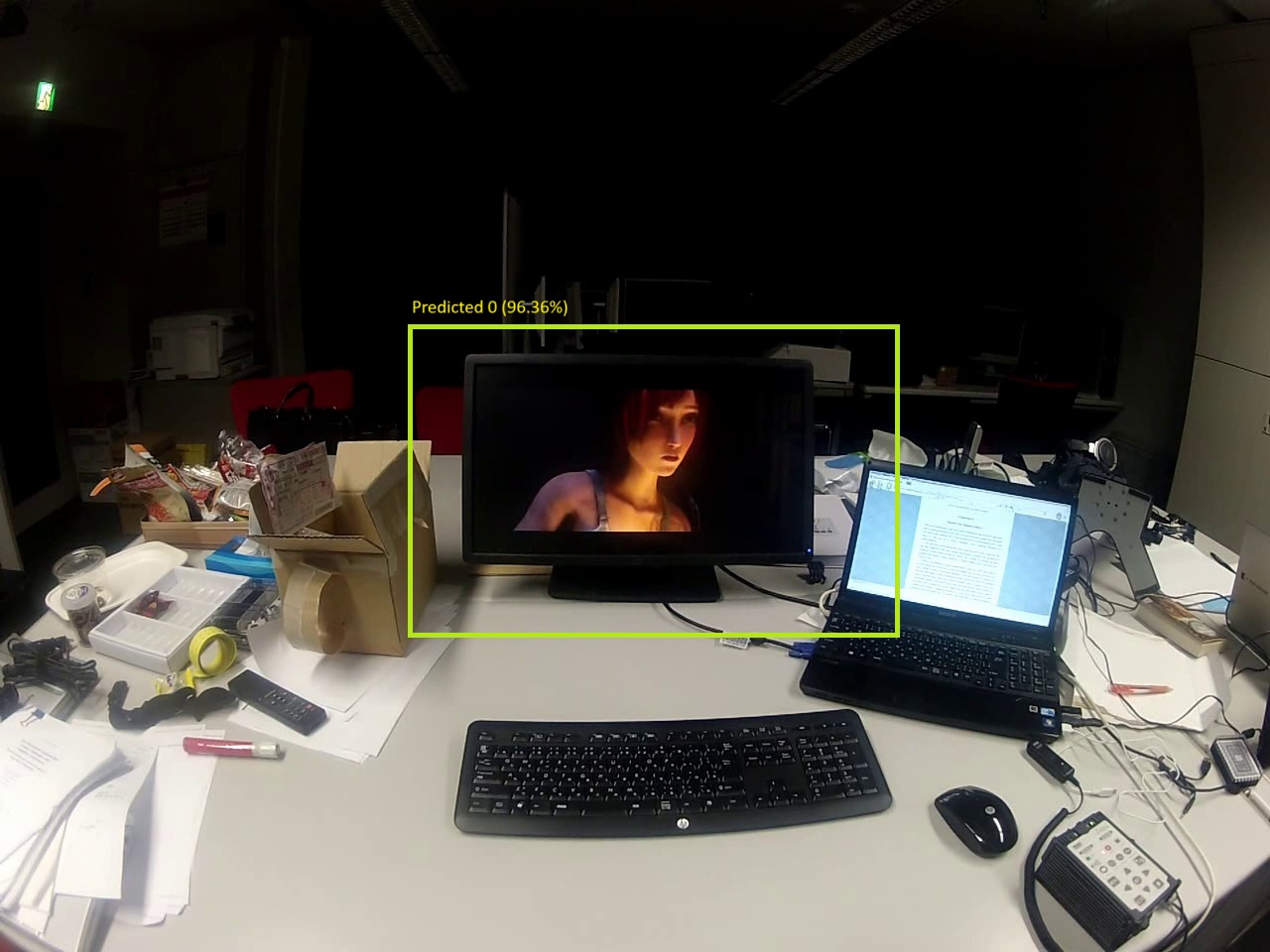}\hfill
    \caption{Video streaming}
  \end{subfigure}
  \hfill
  \begin{subfigure}{0.3\linewidth}
    \includegraphics[width= \linewidth]{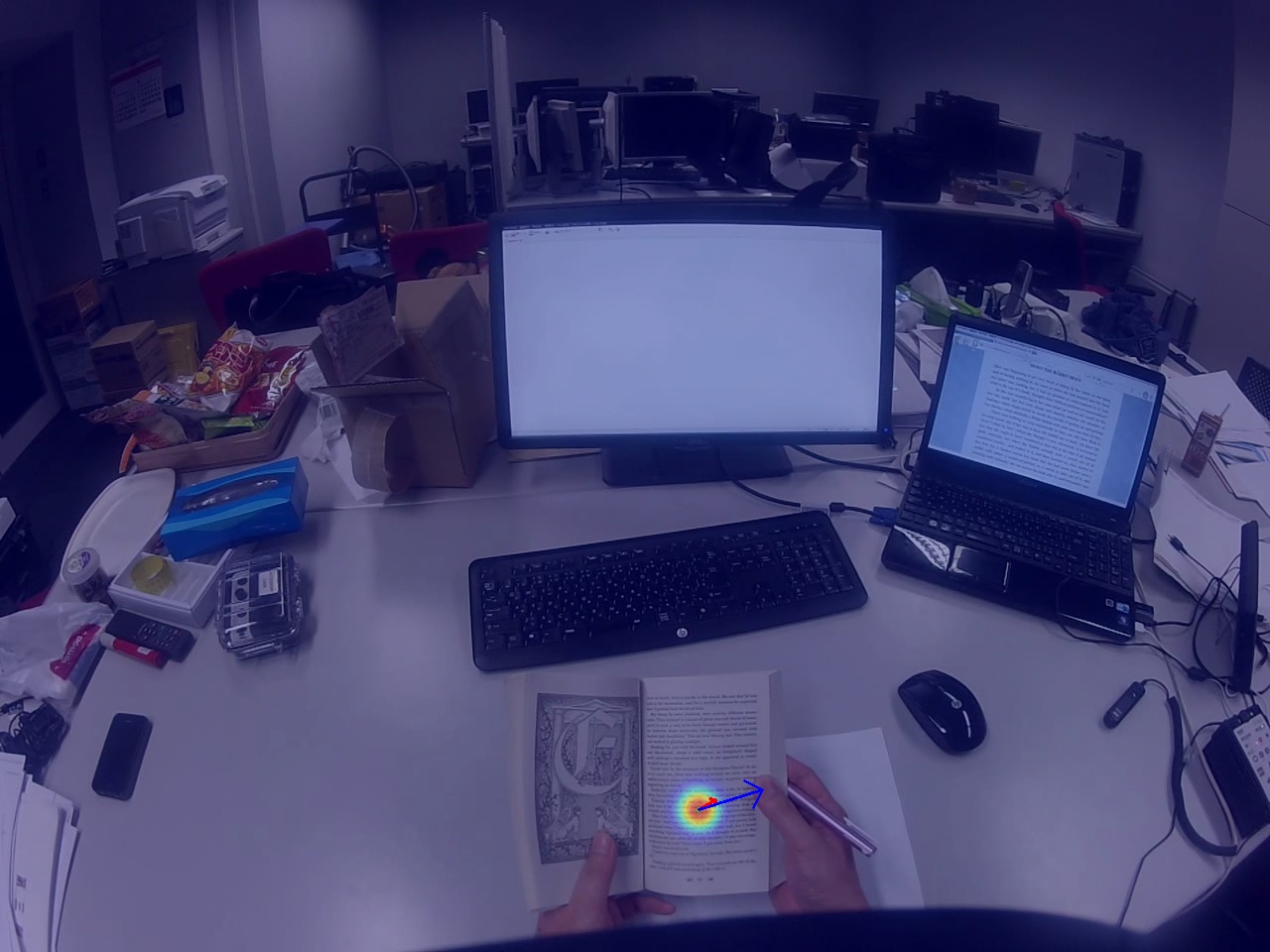}\hfill
    \caption{$d = 0.41$}
  \end{subfigure}
  \hfill
  \begin{subfigure}{0.3\linewidth}
    \includegraphics[width= \linewidth]{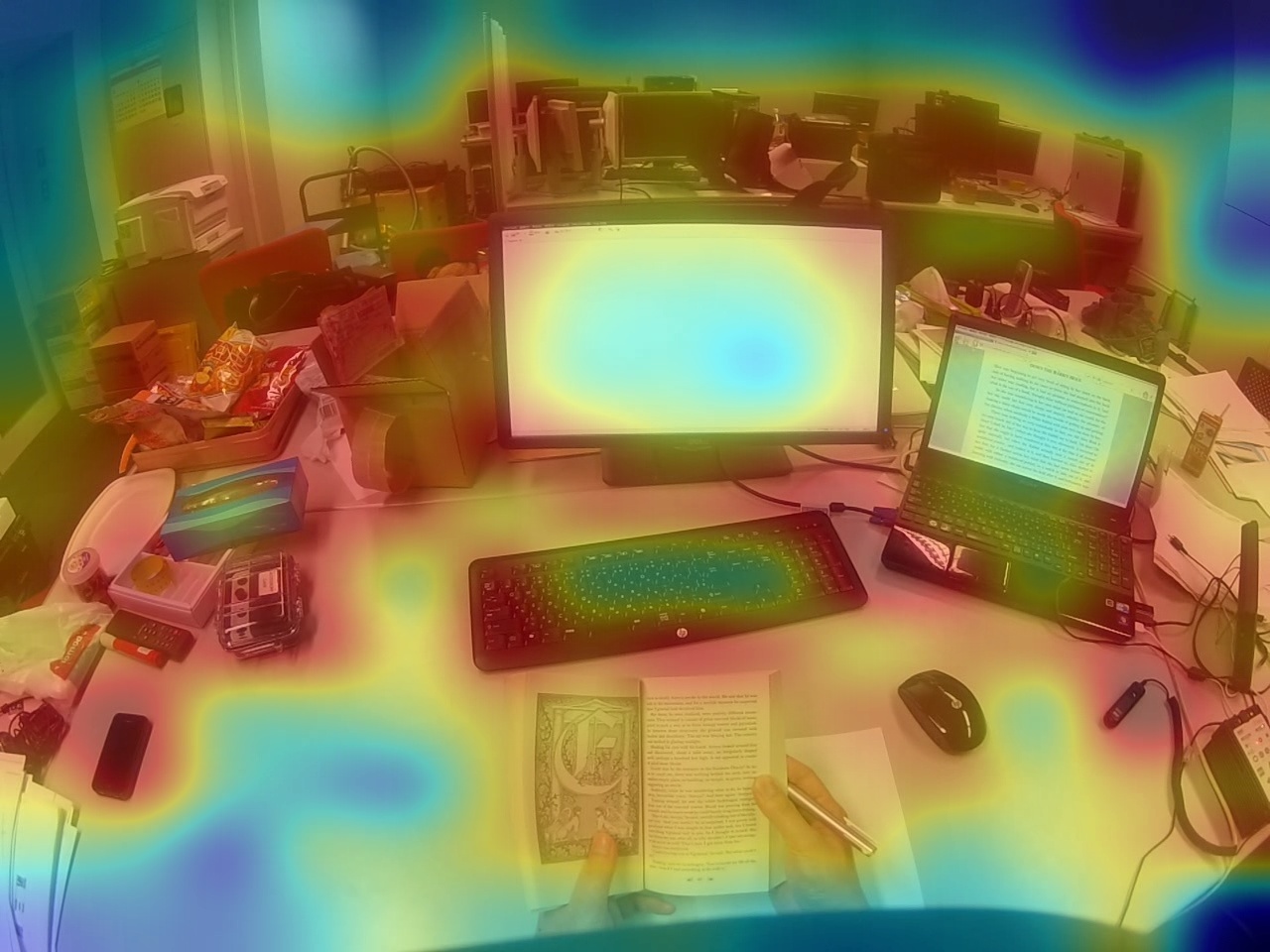}\hfill
    \caption{$p = 0.72$}
  \end{subfigure}
  \hfill
  \begin{subfigure}{0.3\linewidth}
    \includegraphics[width= \linewidth]{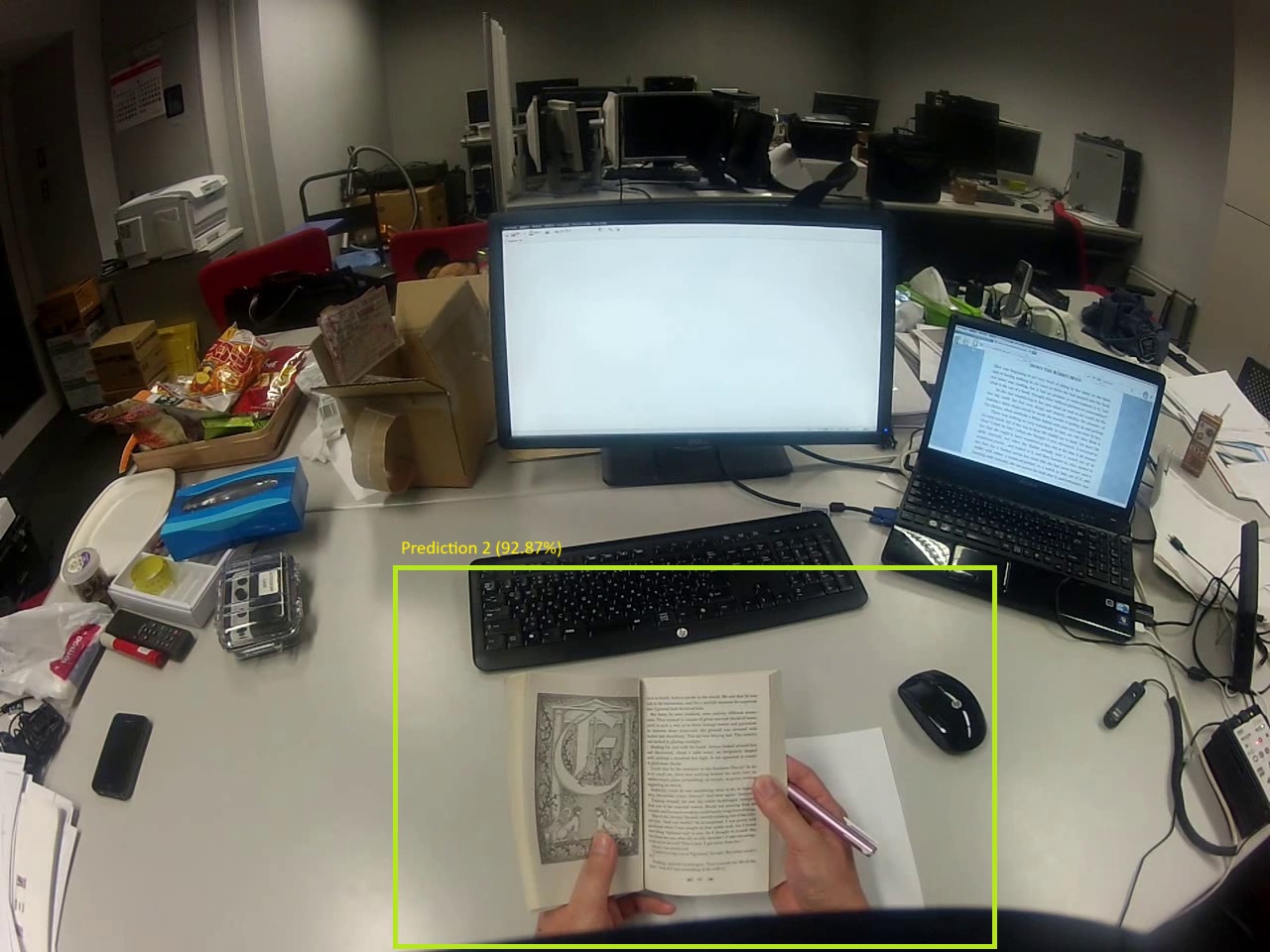}\hfill
    \caption{Reading}
  \end{subfigure}
  \caption{Model predictions for the Ego Motion Dataset during video streaming (a:d) and reading (e:h) tasks, with similar depth $d$ profiles. Bounding box sizes, scaled by pupil dilation $p$, are larger for the reading task, indicating higher visual focus compared to video streaming.}
  \label{fig:results ego motion}
\end{figure}

\begin{figure}[h]
  \centering
  \begin{subfigure}{0.3\linewidth}
    \includegraphics[width= \linewidth]{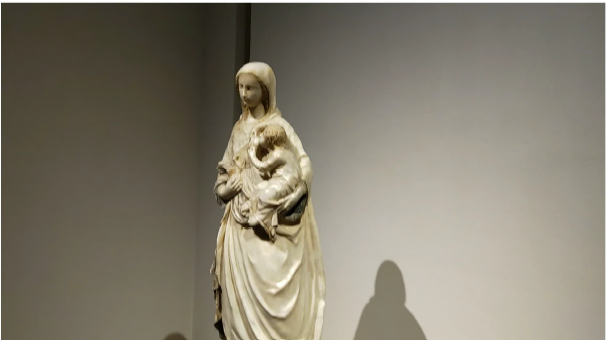}\hfill
    \caption{Original frame}
  \end{subfigure}
  \hfill
  \begin{subfigure}{0.3\linewidth}
    \includegraphics[width= \linewidth]{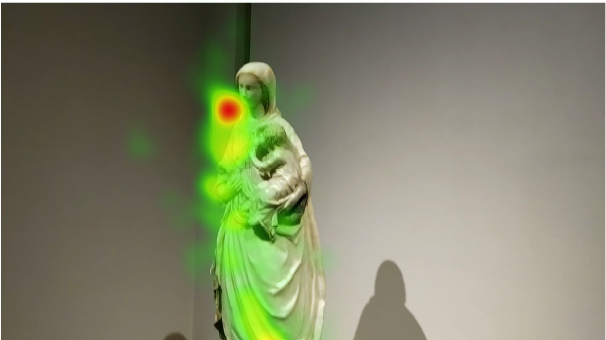}\hfill
    \caption{Attention heatmap}
  \end{subfigure}
  \hfill
  \begin{subfigure}{0.3\linewidth}
    \includegraphics[width= \linewidth]{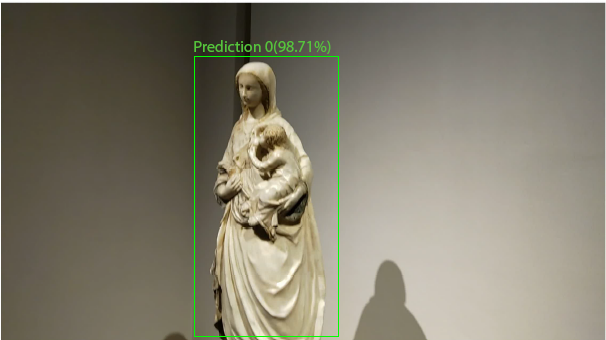}\hfill
    \caption{Madonna del Cardillo}
  \end{subfigure}
  \hfill
  \begin{subfigure}{0.3\linewidth}
    \includegraphics[width= \linewidth]{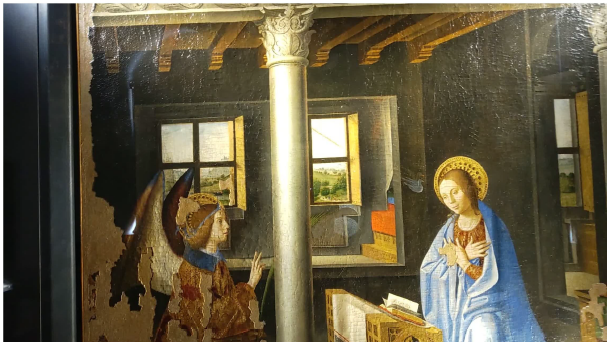}\hfill
    \caption{Original frame}
  \end{subfigure}
  \hfill
  \begin{subfigure}{0.3\linewidth}
    \includegraphics[width= \linewidth]{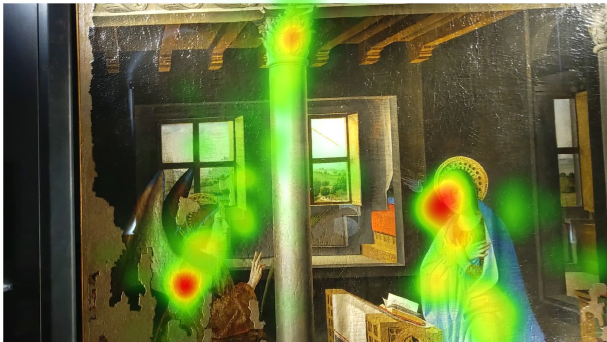}\hfill
    \caption{Attention heatmap}
  \end{subfigure}
  \hfill
  \begin{subfigure}{0.3\linewidth}
    \includegraphics[width= \linewidth]{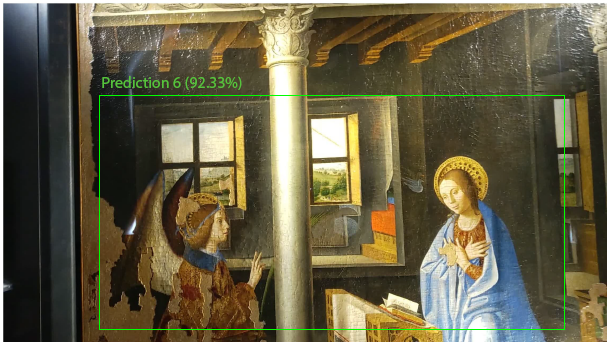}\hfill
    \caption{Annunciazione}
  \end{subfigure}
  \caption{Model predictions for the Ego-CH-Gaze dataset (without depth and pupil dilation information)}
  \label{fig:results ego-ch-gaze}
\end{figure}

Notably, the proposed Eyes on Target model achieves substantial improvements in mAP over the Attended Object of Interest Detection baseline. Its performance is also comparable to the Faster R-CNN benchmark reported in \cite{weaklysupervisedgaze}, with mAP@0.5 of 0.60 and mAP@0.75 of 0.41, values that are closely aligned with our results. This demonstrates that our model provides a stronger alternative to existing gaze-integrated object detection approaches. However, we observe a slight drop in mAP@0.5 and mAP@0.75 metrics against the non-gaze integrated detection models (YOLO v7 and DETR) across both Ego-CH-Gaze and Egocentric Maritime Simulator Dataset. A likely reason is the dynamic scaling of bounding boxes in regions with weaker gaze signals, which reduces overlap with ground truth in overall evaluation. Despite this, extensive qualitative analysis reveals that gaze-guided scaling leads to tighter and more semantically meaningful bounding boxes around areas of visual focus, which validates our localised detection hypothesis for egocentric use cases. \cref{fig:results cems} presents test samples from the Egocentric Maritime Simulator Dataset, showing the model's output across different stages. For closer targets, gaze-modified attention maps exhibit stronger activations near the gaze point. When coupled with larger pupil dilation, this results in broader bounding boxes, with higher IoU, reflecting heightened visual attention. For distant objects, the model distributes attention more sparsely, and bounding boxes are scaled down proportionally, accounting for perspective and depth, thereby maintaining spatial fidelity. Similarly, \cref{fig:results ego motion} illustrates results from the Ego Motion Dataset. Here, bounding box sizes adapt based on pupil diameter, with increased focus observed during reading tasks compared to video watching, yielding slightly larger RoIs. This adaptability highlights the model’s ability to modulate attention and spatial localisation based on task demands and visual depth, enhancing robustness even under subtle variations in scene context. Since the Ego-CH-Gaze dataset does not include depth and pupil dilation information \cref{fig:results ego-ch-gaze} showcases the results of Eyes on Target (gaze position only) model.

Overall, the results validate the effectiveness of our method of integrating gaze and depth signals for object classification and localisation.

\begin{table}[]
\resizebox{0.5\textwidth}{!}{
\begin{tabular}{lllc}
\textbf{Model} & \textbf{Dataset} & \textbf{\begin{tabular}[c]{@{}l@{}}Classification \\ Metrics\end{tabular}} & \multicolumn{1}{l}{\textbf{\begin{tabular}[c]{@{}l@{}}Detection \\ Metrics\end{tabular}}} \\ \hline
\multirow{2}{*}{MoWord+SaWord \cite{egodataset}}  & Ego Motion & \begin{tabular}[c]{@{}l@{}}A = 0.77\\ F1 = 0.76\end{tabular} & - \\ \cline{2-4} 
 & \begin{tabular}[c]{@{}l@{}}Egocentric Maritime \\ Simulator Dataset\end{tabular} & \begin{tabular}[c]{@{}l@{}}A = 0.75\\ F1 = 0.76\end{tabular} & - \\ \hline
 \multirow{2}{*}{{\begin{tabular}[c]{@{}l@{}}Attended Object of \\ Interest Detection \\ \cite{weaklysupervisedgaze}\end{tabular}}  }  & Ego-CH-Gaze & \begin{tabular}[c]{@{}l@{}}A = 0.76\\ F1 = 0.75\end{tabular} & \multicolumn{1}{l}{\begin{tabular}[c]{@{}l@{}}mAP@0.5 = 0.41\\ mAP@0.75 = 0.19\end{tabular}} \\ \cline{2-4} 
 & \begin{tabular}[c]{@{}l@{}}Egocentric Maritime \\ Simulator Dataset\end{tabular} & \begin{tabular}[c]{@{}l@{}}A = 0.50\\ F1 = 0.49\end{tabular} & \multicolumn{1}{l}{\begin{tabular}[c]{@{}l@{}}mAP@0.5 = 0.37\\ mAP@0.75 = 0.13\end{tabular}} \\ \hline
\multirow{3}{*}{Eyes On Target} & Ego Motion & \textbf{\begin{tabular}[c]{@{}l@{}}A = 0.94\\ F1 = 0.94\end{tabular}} & - \\ \cline{2-4} 
& Ego-CH-Gaze & \textbf{\begin{tabular}[c]{@{}l@{}}A = 0.89\\ F1 = 0.90\end{tabular}} & \multicolumn{1}{l}{\begin{tabular}[c]{@{}l@{}}mAP@0.5 = 0.58\\ mAP@0.75 = 0.37\end{tabular}} \\\cline{2-4} 
 & \begin{tabular}[c]{@{}l@{}}Egocentric Maritime \\ Simulator Dataset\end{tabular} & \textbf{\begin{tabular}[c]{@{}l@{}}A = 0.92\\ F1 = 0.91\end{tabular}} & \multicolumn{1}{l}{\begin{tabular}[c]{@{}l@{}}mAP@0.5 = 0.61\\ mAP@0.75 = 0.55\end{tabular}} \\\hline
\multirow{3}{*}{\begin{tabular}[c]{@{}l@{}}YOLO v7 \\ Object Detection\end{tabular}} & Ego Motion & \begin{tabular}[c]{@{}l@{}}A = 0.93\\ F1 = 0.93\end{tabular} & - \\ \cline{2-4} 
& Ego-CH-Gaze & \begin{tabular}[c]{@{}l@{}}A = 0.87\\ F1 = 0.88\end{tabular} & \multicolumn{1}{l}{\begin{tabular}[c]{@{}l@{}}mAP@0.5 = 0.6\\ mAP@0.75 = 0.42\end{tabular}} \\\cline{2-4}
 & \begin{tabular}[c]{@{}l@{}}Egocentric Maritime \\ Simulator Dataset\end{tabular} & \begin{tabular}[c]{@{}l@{}}A = 0.84\\ F1 = 0.83\end{tabular} & \multicolumn{1}{l}{\begin{tabular}[c]{@{}l@{}}mAP@0.5 = 0.66\\ mAP@0.75 = 0.61\end{tabular}} \\ \hline
\multirow{2}{*}{DETR} & Ego Motion & \begin{tabular}[c]{@{}l@{}}A = 0.92\\ F1 = 0.93\end{tabular} & - \\ \cline{2-4}
& Ego-CH-Gaze & \begin{tabular}[c]{@{}l@{}}A = 0.86\\ F1 = 0.86\end{tabular} & \multicolumn{1}{l}{\begin{tabular}[c]{@{}l@{}}mAP@0.5 = 0.59\\ mAP@0.75 = 0.41\end{tabular}} \\\cline{2-4}
 & \begin{tabular}[c]{@{}l@{}}Egocentric Maritime \\ Simulator Dataset\end{tabular} & \begin{tabular}[c]{@{}l@{}}A = 0.89\\ F1 = 0.87\end{tabular} & \multicolumn{1}{l}{\begin{tabular}[c]{@{}l@{}}mAP@0.5 = 0.64\\ mAP@0.75 = 0.60\end{tabular}} \\ 
\end{tabular}}
\caption{Comparison of classification metrics from contemporary models on Ego Motion Dataset, Ego-CH-Gaze Dataset, and Egocentric Maritime Simulator Dataset}
\label{tab:classification metrics}
\end{table}

\subsection{Ablation Study}
\Cref{tab:ablation study results} shows the impact of individual gaze components on classification accuracy for the Egocentric Maritime Simulator Dataset. Incorporating gaze direction $\hat{g}$ significantly improves attention localisation, highlighting the insufficiency of gaze points alone. While pupil dilation $p$ and depth $d$ do not directly modify attention, they enhance bounding box scaling, improving classification accuracy. 

\begin{table}
\centering
\resizebox{0.4\textwidth}{!}{
\begin{tabular}{lll}
\textbf{Model} & \textbf{\begin{tabular}[c]{@{}l@{}}Egocentric Maritime \\ Simulator Dataset\end{tabular}} \\ \hline

\raggedright{No gaze component}                             & 0.891 \\
\raggedright{Including $g_{(x,y)}$}                         & 0.909 \\
\raggedright{Including $g_{(x,y)}$ and $\hat{g}$}           & 0.917 \\
\raggedright{Including $g_{(x,y)}$, $\hat{g}$, $p$ and $d$} & 0.924
\end{tabular}}
\caption{Ablation from standard ViT model(DETR) to Eyes on Target on Classification Accuracy}
\label{tab:ablation study results}
\end{table}

To evaluate the impact of $\lambda_1$, $\lambda_2$, and $\lambda_3$ on the adjustment of the bounding box size, we conducted experiments with different combinations of these parameters. Our solution focuses on localised classification; as a result, even if the bounding boxes are not perfectly tight, allowing some object portions to fall outside or leaving excess space, accurate classification is still achievable, which is the motivation to optimise for classification accuracy over IoU. \cref{tab:bounding box scaling} highlights the results on the Egocentric Maritime Simulator Dataset test set, indicating that depth ($\lambda_3$) significantly influences classification accuracy more than visual focus ($\lambda_2$), and while $\text{AttnScore}$ ($\lambda_1$) improves global attention around the gaze point, it does not substantially affect precise localisation.

\begin{table}
\centering
\resizebox{0.5\textwidth}{!}{
\begin{tabular}{lll}
\textbf{$\lambda_1, \lambda_2, \lambda_3$} & \textbf{\begin{tabular}[c]{@{}l@{}}Classification\\ Accuracy\end{tabular}} & \textbf{\begin{tabular}[c]{@{}l@{}}Mean IoU with\\ object in gaze\end{tabular}} \\ \hline
$\lambda_1=0.5, \lambda_2=0.2, \lambda_3=0.2$ & 0.909 & 0.80 \\
$\lambda_1=0.5, \lambda_2=0.2, \lambda_3=0.3$ & 0.919 & 0.82 \\
$\lambda_1=0.5, \lambda_2=0.2, \lambda_3=0.5$ & 0.921 & 0.82 \\
$\lambda_1=0.5, \lambda_2=0.3, \lambda_3=0.5$ & 0.927 & 0.82 \\
$\lambda_1=0.5, \lambda_2=0.5, \lambda_3=0.5$ & 0.922 & 0.82 \\
$\lambda_1=0.7, \lambda_2=0.3, \lambda_3=0.5$ & 0.909 & 0.81 \\
$\lambda_1=0.3, \lambda_2=0.3, \lambda_3=0.5$ & 0.924 & 0.80
\end{tabular}}
\caption{Analysis of the influence of RoI scaling parameters $\lambda_1, \lambda_2, \lambda_3$ on IoU measured against annotated test set and classification accuracy}
\label{tab:bounding box scaling}
\end{table}

To evaluate the impact of $\text{GazeBias(i, j)}$ on attention (\cref{eq modified attention}) and model performance, experiments were conducted on the Egocentric Maritime Simulator Dataset with varying $\alpha$ values (\cref{tab:gaze information variation}). The results show that IoU improves steadily with increased $\alpha$, improving attention location. Classification accuracy peaks at $\alpha = 0.7$, suggesting that this level of overlap between model attention and gaze regions provides the most reliable classification. While attention alignment—measured by cosine similarity with the gaze region—continues to improve as $\alpha$ increases, this comes at the cost of classification performance. Beyond $\alpha = 0.7$, attention becomes overly concentrated on the gaze point, reducing the model’s ability to leverage broader image features essential for accurate prediction.

\begin{table}
\centering
\resizebox{0.35\textwidth}{!}{
\begin{tabular}{llll}
\textbf{$\alpha$} & \textbf{\begin{tabular}[c]{@{}l@{}}Classification\\ Accuracy\end{tabular}} & \textbf{\begin{tabular}[c]{@{}l@{}}Attention Alignment \\ (Cosine Similarity)\end{tabular}} \\ 
\hline
0.0 & 0.902 & 0.72 \\
0.1 & 0.906 & 0.77 \\
0.2 & 0.907 & 0.78 \\
0.5 & 0.922 & 0.81 \\
0.7 & 0.924 & 0.89 \\
1.0 & 0.921 & 0.90 \\
2.0 & 0.916 & 0.92 
\end{tabular}}
\caption{Analysis of the influence of gaze information by varying $\alpha$ values}
\label{tab:gaze information variation}
\end{table}

\subsection{Gaze-Aware Head Importance Metrics}

The variation in head importance does not directly correlate with performance improvements but provides valuable interpretability into how gaze information affects attention allocation. As shown in \cref{tab:head importance metric}, we examine three sample attention heads, $I^{prob}_{1}$, $I^{prob}_{5}$, and $I^{prob}_{6}$ (the 1st, 5th, and 6th heads from the second attention layer), comparing their relative importance in the baseline DETR model versus the proposed Eyes on Target model on the Egocentric Maritime Simulator Dataset. The inclusion of gaze data significantly alters the distribution of head importance across these layers.

\begin{figure}
  \centering
  \begin{subfigure}{0.48\linewidth}
    \includegraphics[width= \linewidth]{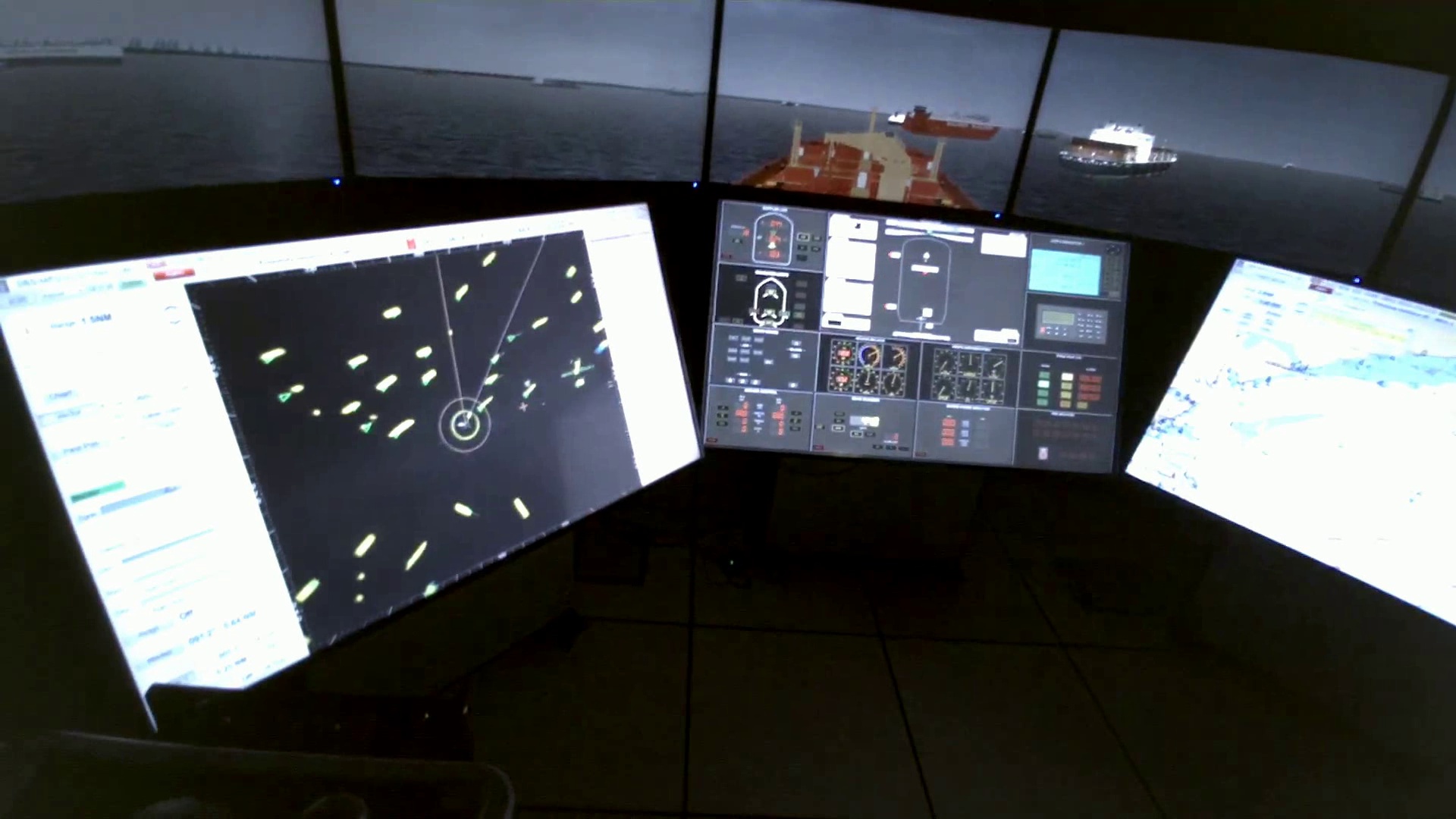}\hfill
    \caption{}
    \label{fig:original}
  \end{subfigure}
  \begin{subfigure}{0.48\linewidth}
    \includegraphics[width= \linewidth]{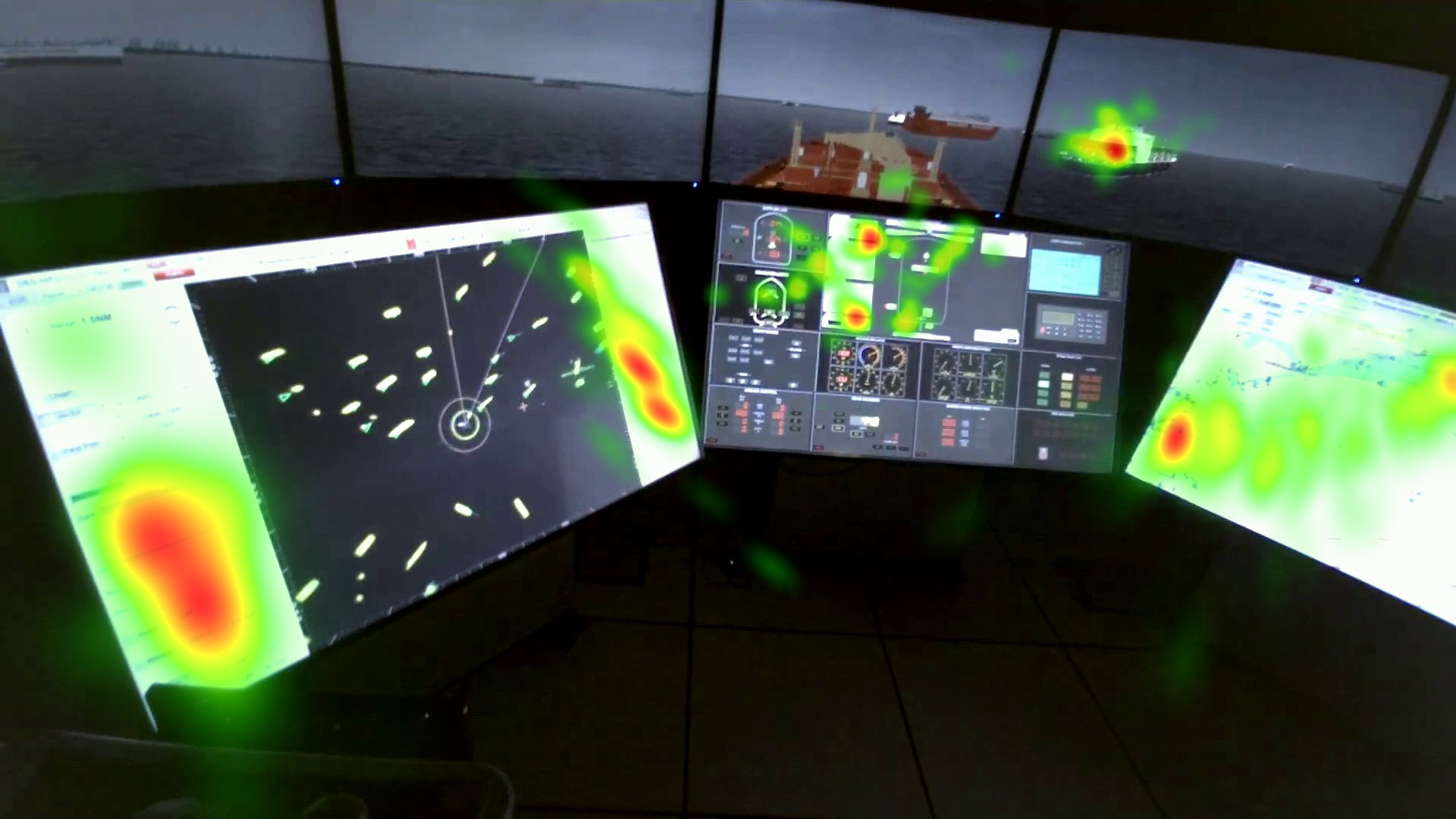}\hfill
    \caption{}
    \label{fig:attention heads}
  \end{subfigure}
  \vfill
  \begin{subfigure}{0.48\linewidth}
    \includegraphics[width= \linewidth]{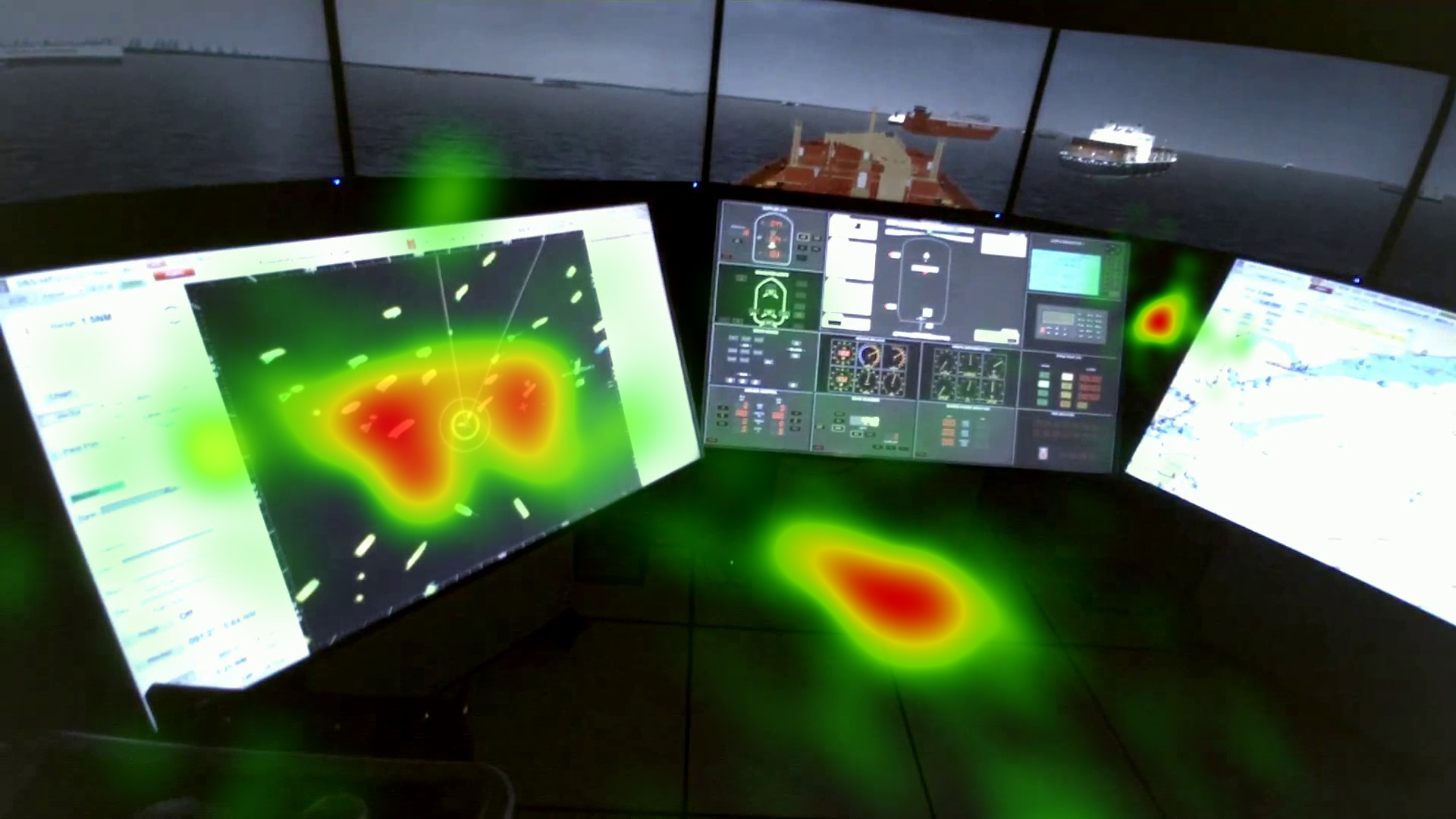}\hfill
    \caption{}
    \label{fig:attention heads}
  \end{subfigure}
  \begin{subfigure}{0.48\linewidth}
    \includegraphics[width= \linewidth]{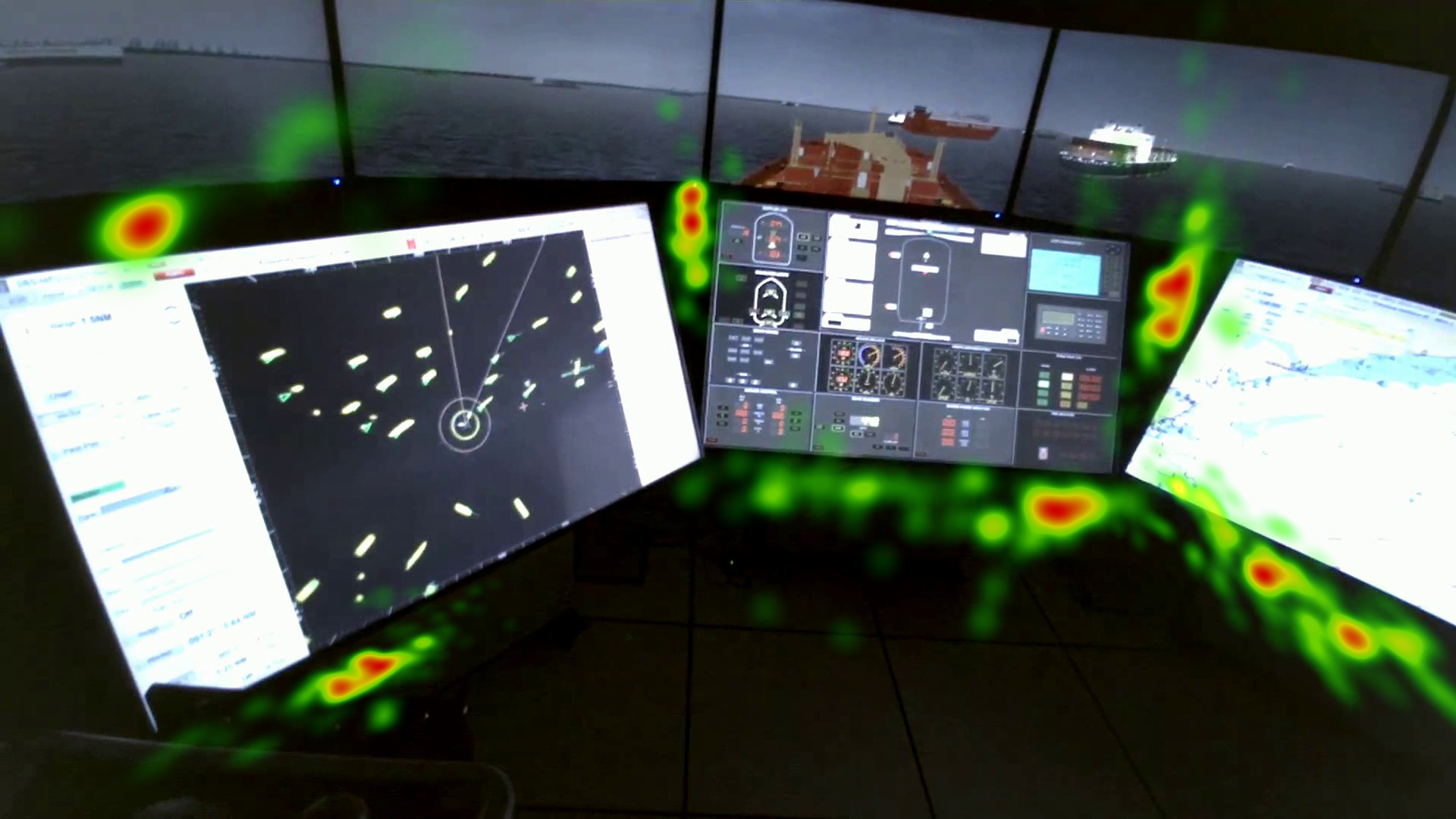}\hfill
    \caption{}
    \label{fig:attention heads}
  \end{subfigure}
  \caption{Qualitative visualization of attention map (\textgreater60\%) from heads $1^{st}$ (red), $5^{th}$(blue) and $6^{th}$(green) from Layer 2 DETR}
  \label{fig:attention heads visualisation}
\end{figure}

Qualitative visualisations in \cref{fig:attention heads visualisation} reveal that the $1^{st}$ and $5^{th}$ heads predominantly respond to brightness variations, while the $6^{th}$ head is more edge-sensitive, based on intensity maps plotted at 60\% or higher attention scores. Interestingly, the increased prominence of the $1^{st}$ and $5^{th}$ heads in the gaze-integrated model may correspond to their alignment with typical gaze fixation points, particularly on interface elements like radar and control panels during simulator exercises. Overall, these findings suggest that gaze information reshapes the internal attention dynamics of the model, selectively amplifying the relevance of heads whose focus aligns with human visual behaviour. This modulation is influenced by spatial patch position, gaze location, and gaze direction, illustrating the nuanced role of gaze in steering both attention distribution and feature representation.

\begin{table}[]
\centering
\begin{tabular}{llll}
\textbf{Model} & \textbf{$I^{prob}_{1}$} & \textbf{$I^{prob}_{5}$} & \textbf{$I^{prob}_{6}$} \\ \toprule
DETR & 0.56 & 0.41 & 0.65 \\
Eyes on Target & 0.68 & 0.54 & 0.62
\end{tabular}
\caption{Changes in attention head importance with and without gaze information integration}
\label{tab:head importance metric}
\end{table}

\section{Conclusion}
This work demonstrates the effectiveness of integrating human gaze cues, including gaze point, depth, pupil dilation, and direction into Vision Transformer architectures for improved object detection in egocentric video scenarios. By aligning model attention with human visual focus, we show that gaze-guided attention mechanisms can substantially enhance performance, especially in tasks where context and intent play a crucial role. Our proposed model, Eyes on Target, introduces a depth-aware, gaze-biased attention formulation within a ViT backbone, enabling fine-grained localisation and robust classification in egocentric perspectives. Evaluations on the Egocentric Maritime Simulator Dataset show that our method outperforms both standard ViT models and state-of-the-art object detectors like YOLO v7 in classification accuracy. On the Ego Motion and Ego-CH-Gaze datasets, our approach also achieves superior results compared to prior gaze-based methods.

Furthermore, we propose a novel gaze-aware head importance metric to quantify how gaze features influence the internal attention dynamics of transformer heads. This analytical tool provides interpretable insights into which components of the model are most sensitive to human visual patterns, offering a foundation for future work in explainable attention and human-in-the-loop learning.

Overall, this research further explores and highlights the potential of gaze signals in vision models for object classification and detection and sets the stage for further integration of human attention into deep learning frameworks, particularly in safety-critical and simulation-based training environments with an egocentric perspective.

\bibliographystyle{IEEEbib}
\bibliography{refs}

\end{document}